\documentclass[runningheads]{llncs}

\usepackage{eccv}

\usepackage{eccvabbrv}

\usepackage{graphicx}
\usepackage{booktabs}
\usepackage{microtype}

\usepackage[accsupp]{axessibility}  % Improves PDF readability for those with disabilities.

\usepackage{hyperref}

\usepackage{orcidlink}

\usepackage{color}
\usepackage{soul}
\usepackage{tabularx}
\usepackage{mathtools}
\usepackage{algpseudocode}
\usepackage{algorithm}
\usepackage{enumitem}

\usepackage[font=small]{caption}

\usepackage[dvipsnames]{xcolor}
\usepackage{colortbl}

\newcommand{\E}{\mathbb{E}}
\newcommand{\R}{\mathbb{R}}
\newcommand{\LL}{\mathcal{L}}
\newcommand{\NN}{\mathsf{N}}
\newcommand{\rd}{{\mathrm{d}}} 
\newcommand{\bx}{{\mathbf{x}}}
\newcommand{\bX}{{\mathbf{X}}}
\newcommand{\by}{{\mathbf{y}}}
\newcommand{\bv}{{\mathbf{v}}}
\newcommand{\bs}{{\mathbf{s}}}
\newcommand{\bW}{{\mathbf{W}}}
\newcommand{\bk}{{\mathbf{k}}}

\newcommand{\eps}{{\boldsymbol \varepsilon}}

\definecolor{ForestGreen}{RGB}{34,139,34}

\makeatletter
\def\@fnsymbol#1{\ensuremath{\ifcase#1\or \dagger\or \ddagger\or
   \mathsection\or \mathparagraph\or \|\or **\or \dagger\dagger
   \or \ddagger\ddagger \else\@ctrerr\fi}}
    \makeatother

\newcommand{\forest}[1]{\textcolor{ForestGreen}{#1}}
\newcommand{\violet}[1]{\textcolor{violet}{#1}}
\newcommand{\orange}[1]{\textcolor{orange}{#1}}
\newcommand{\teal}[1]{\textcolor{teal}{#1}}

\begin{document}

% ---------------------------------------------------------------
% TODO REVIEW: Replace with your title
\title{SiT: Exploring Flow and Diffusion-based Generative Models with Scalable Interpolant Transformers} 

% TODO REVIEW: If the paper title is too long for the running head, you can set
% an abbreviated paper title here. If not, comment out.
\titlerunning{Scalable Interpolant Transformers}

% TODO FINAL: Replace with your author list. 
% Include the authors' OCRID for the camera-ready version, if at all possible.
\author{Nanye Ma \and
Mark Goldstein \and
Michael S. Albergo \and
Nicholas M. Boffi \and
Eric Vanden-Eijnden\thanks{Equal advising.} \and
Saining Xie$^\dagger$}

% TODO FINAL: Replace with an abbreviated list of authors.
\authorrunning{N.Ma et al.}
% First names are abbreviated in the running head.
% If there are more than two authors, 'et al.' is used.

% TODO FINAL: Replace with your institution list.
\institute{New York University}

\maketitle

\begin{abstract}
We present Scalable Interpolant Transformers (SiT), a family of generative models built on the backbone of Diffusion Transformers (DiT). The interpolant framework, which allows for connecting two distributions in a more flexible way than standard diffusion models, makes possible a modular study of various design choices impacting generative models built on dynamical transport: learning in discrete or continuous time, the objective function, the interpolant that connects the distributions, and deterministic or stochastic sampling.
By carefully introducing the above ingredients, SiT surpasses DiT uniformly across model sizes on the conditional ImageNet $256 \times 256$ and $512 \times 512$ benchmark using the exact same model structure, number of parameters, and GFLOPs. By exploring various diffusion coefficients, which can be tuned separately from learning, SiT achieves an FID-50K score of 2.06 and 2.62, respectively. 
Code is available here: \href{https://github.com/willisma/SiT}{https://github.com/willisma/SiT}
\end{abstract}

%%%%%%%%%%%%%%%%%%%%%%%%%%%%%%%%%%%%%%%%%%%%%%%%%%%%%%%%%%%%%
%%%%%%%%%%%%%%%%%%%%%%%%%%% INTRO %%%%%%%%%%%%%%%%%%%%%%%%%%%
%%%%%%%%%%%%%%%%%%%%%%%%%%%%%%%%%%%%%%%%%%%%%%%%%%%%%%%%%%%%%

\section{Introduction}
\label{sec:intro}

Contemporary success in image generation has come from a combination of algorithmic advances, improvements in model architecture, and progress in scaling neural network models and data.
State-of-the-art diffusion models~\cite{ho2020denoising, rombach2022high} proceed by incrementally transforming data into Gaussian noise as prescribed by an iterative stochastic process, which can be specified either in discrete or continuous time. 
At an abstract level, this corruption process can be viewed as defining a time-dependent distribution that is iteratively smoothed from the original data distribution into a standard normal distribution.
Diffusion models learn to reverse this corruption process and push Gaussian noise backwards along this connection to obtain data samples. 
The objects learned to perform this transformation conventionally predict either the noise in the corruption process~\cite{ho2020denoising} or the score of the distribution that connects the data and the Gaussian~\cite{song2021scorebased}, though alternatives of these choices exist
\cite{salimans2022progressive, hoogeboom2023simple}.
\begin{table}
  \centering
  \caption{\textbf{Scalable Interpolant Transformers.} We systematically vary the following aspects of a generative model: \textbf{\forest{time discretization}}, \textbf{\orange{model prediction}}, \textbf{\violet{interpolant}}, and \textbf{\teal{sampler}}. The resulting Scalable Interpolant Transformer (SiT) model, under identical training compute, consistently outperforms the Diffusion Transformer (DiT) in generating 256×256 ImageNet images. All models employ a patch size of 2. In this work, we ask the question: \textit{What is the source of the performance gain?}}
  \scalebox{0.9}{
  \begin{tabular}[ht]{lccc}
    \toprule
    Model & Params(M) & Training Steps & FID $\downarrow$\\
    \midrule
    DiT-S & 33 & 400K & 68.4 \\
    SiT-S & 33 & 400K & \textbf{57.6} \\
    \midrule
    DiT-B & 130 & 400K & 43.5\\
    SiT-B & 130 & 400K & 
    \textbf{33.0}\\
    \midrule
    DiT-L & 458 & 400K & 23.3\\
    SiT-L & 458 & 400K & 
    \textbf{18.8} \\  
    \midrule
    DiT-XL & 675 & 400K & 19.5 \\
    SiT-XL & 675 & 400K & \textbf{17.2} \\
    \midrule
    DiT-XL & 675 & 7M & 9.6 \\
    SiT-XL & 675 & 7M & \textbf{8.3} \\
    \midrule
    DiT-XL\ $_{\text{(cfg=1.5)}}$ & 675 & 7M  & 2.27 \\
    SiT-XL\ $_{\text{(cfg=1.5)}}$ & 675 & 7M  & \textbf{2.06} \\
    \bottomrule
  \end{tabular}
  }
  \label{tab:FID-50K}
  % \vspace{-2em}
\end{table}
\begin{figure*}[h]
  \centering
  
   \includegraphics[width=0.8\linewidth]{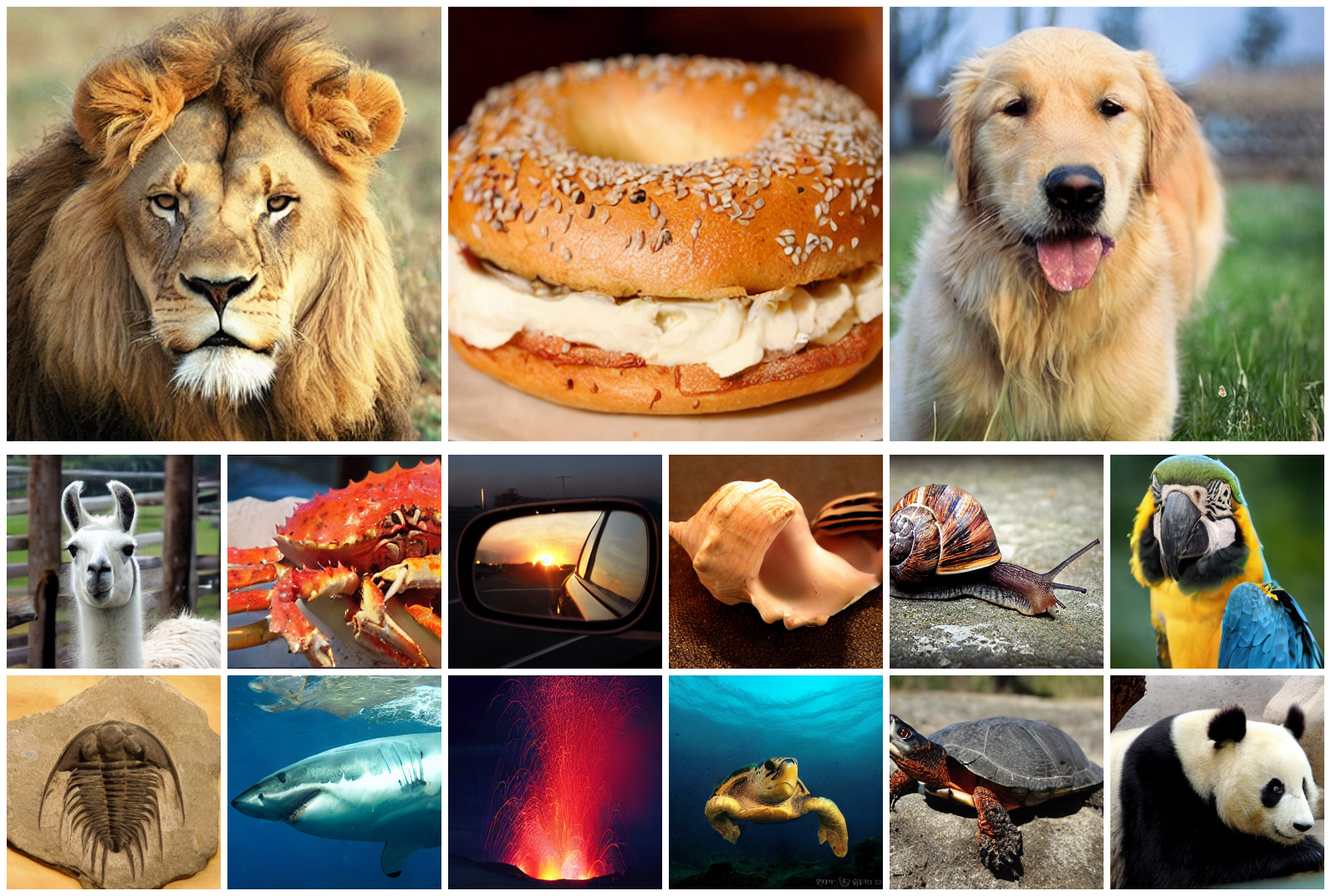}
   % \vspace{-0.5em}
    \caption{Selected samples from SiT-XL models trained on ImageNet~\cite{russakovsky2015imagenet} at $512\times512$ and $256\times256$ resolution with cfg = 4.0, respectively.}
   \label{fig:sota}
   % \vspace{-1em}
\end{figure*}
\begin{figure*}[ht!]
  \centering
  \includegraphics[width=0.24\linewidth]{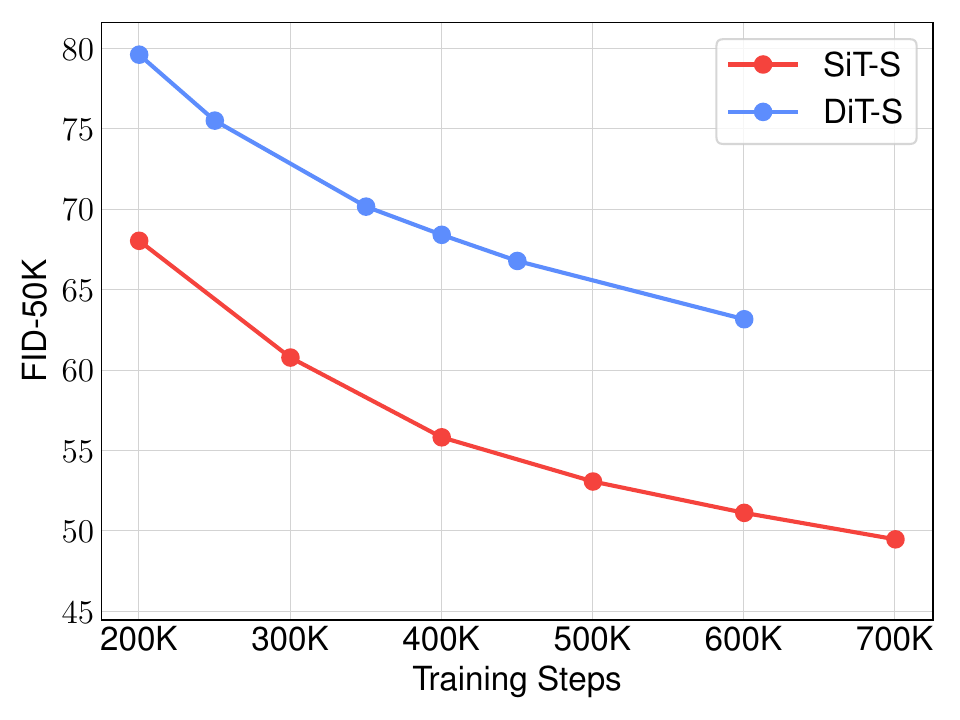}
   \includegraphics[width=0.24\linewidth]{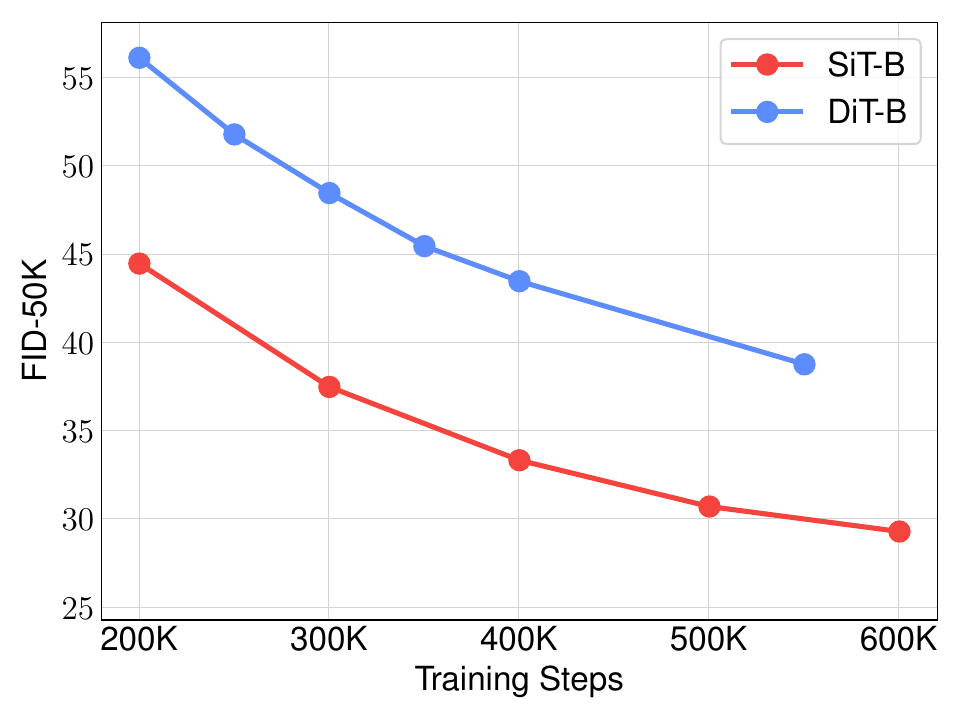}
   \includegraphics[width=0.24\linewidth]{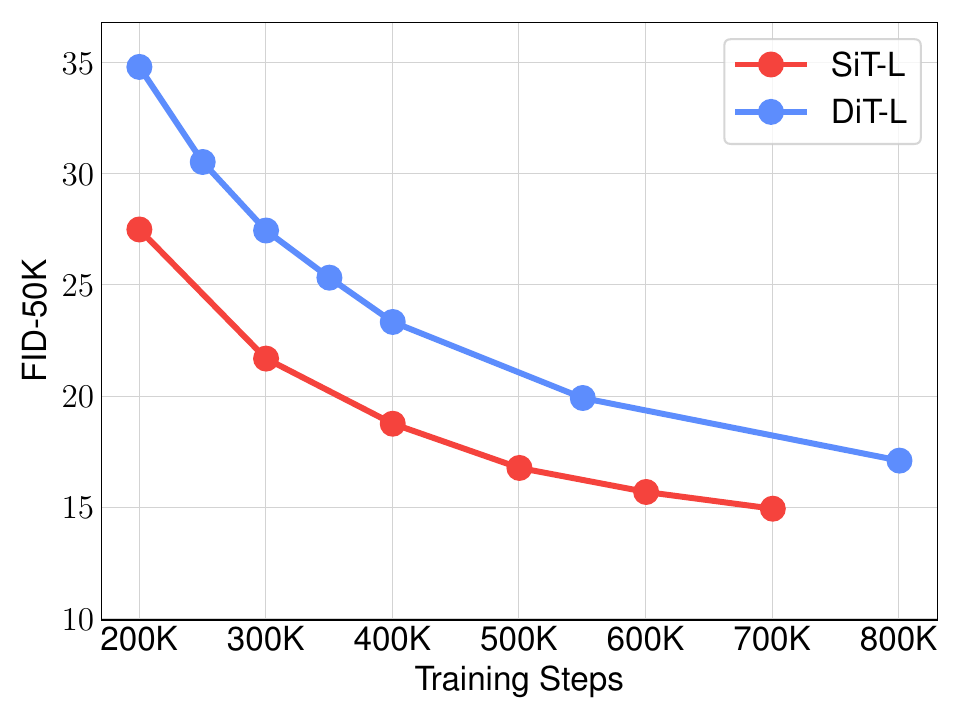}
   \includegraphics[width=0.24\linewidth]{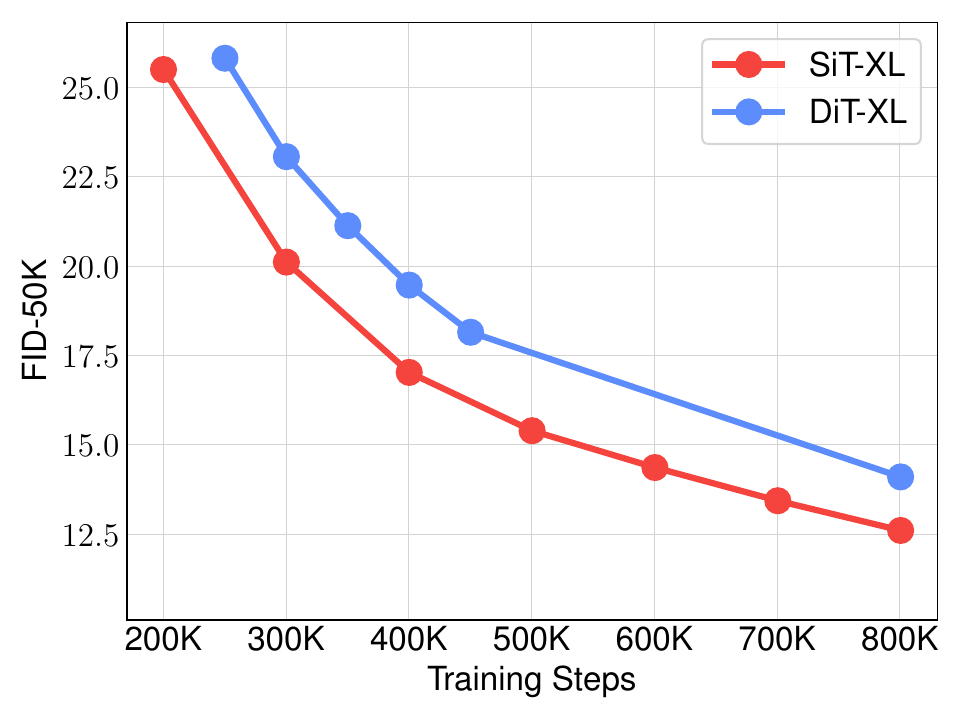}
    % \vspace{-0.5em}
   \caption{\textbf{SiT improves FID across all model sizes.} FID-50K over training iterations for both DiT and SiT. All results are produced by a Euler-Maruyama sampler using 250 integration steps. Across all model sizes, SiT converges much faster.
   }
   % \vspace{-2em}
   \label{fig:scale_sampler}
\end{figure*}
While diffusion models originally represented these objects with a U-Net architecture~\cite{ho2020denoising, ronneberger2015}, recent work has highlighted that architectural advances in vision such as the Vision Transformer (ViT)~\cite{dosovitskiy2021image} can be incorporated into the standard diffusion model pipeline to improve performance~\cite{peebles2023scalable}. 

Orthogonally, significant research effort has gone into exploring the structure of the noising process, which has been shown to lead to performance benefits~\cite{kingma2023variational, karras2022elucidating, kingma2023understanding, singhal2023diffuse}.
Yet, many of these efforts do not move past the notion of passing data through a diffusion process with an equilibrium distribution, which is a restricted type of connection between the data and the Gaussian.
Recently-introduced \textit{stochastic interpolants}~\cite{albergo2023stochastic} lift this constraint and introduce more flexibility in the noise-data connection. 
In this paper, we systematically explore the effect of this flexibility on performance in large scale image generation.

Intuitively, we expect that the difficulty of the \textit{learning problem} can be related to both the specific connection chosen and the object that is learned. 
Our aim is to clarify these design choices, so as to simplify the learning problem and thereby improve performance. 
To understand where potential benefits arise in the learning problem, we start with Denoising Diffusion Probabilistic Models (DDPMs) and sweep through adaptations of: (i) which object to learn, and (ii) which interpolant to choose to reveal best practices. 

In addition to the learning problem, there is a \textit{sampling problem} that must be solved at inference time.
It has been acknowledged for diffusion models that sampling can be either deterministic or stochastic \cite{song2021maximum}, and the choice of sampling method can be made after the learning process.
Yet, the diffusion coefficients used for stochastic sampling are typically presented as intrinsically tied to the forward noising process, which need not be the case in general.

Throughout this paper, we explore how the design of the interpolant and the use of the resulting model as either a deterministic or a stochastic sampler impact performance. 
We gradually transition from a typical denoising diffusion model to an interpolant model by taking a series of orthogonal steps in the design space.
As we progress, we carefully evaluate how each move away from  the diffusion model impacts the performance.
In summary, our \textbf{main contributions} are:

\begin{itemize}
    \item We systematically study the SiT design space through the combinations of the four key components: \textbf{\forest{time discretization}}, \textbf{\orange{model prediction}}, \textbf{\violet{interpolant}}, and \textbf{\teal{sampler}}. 
    \item We provide theoretical motivation for the choice of each component and study how they lead to improved practical performance.
    \item We exploit the tunability of the diffusion coefficient of the stochastic sampler, and show that its adaptation can tighten control of the KL-divergence between the model and the target. We show how this leads to empirical benefits without any additional re-training.
    \item Combining the best design choices identified in each component, our SiT model surpasses Diffusion Transformer(DiT) on both $256 \times 256$ and $512 \times 512$ image resolution, achieving FID-50K scores of 2.06 and 2.62, respectively, without modifying any structure or hyperparameter of the model.
\end{itemize}

%%%%%%%%%%%%%%%%%%%%%%%%%%%%%%%%%%%%%%%%%%%%%%%%%%%%%%%%%%%%%
%%%%%%%%%%%%%%%%%%%%%%%%%%%% SiT %%%%%%%%%%%%%%%%%%%%%%%%%%%%
%%%%%%%%%%%%%%%%%%%%%%%%%%%%%%%%%%%%%%%%%%%%%%%%%%%%%%%%%%%%%

\section{SiT: Scalable Interpolant Transformers}
\label{sec:sit}

We begin by recalling the main ingredients for building flow-based and diffusion-based generative models.

\subsection{Flows and diffusions}
\label{sec:generative}
Flow and diffusion models both utilize stochastic processes to gradually turn noise $\eps \sim \NN(0, \mathbf{I})$ into data $\mathbf{x}_* \sim p(\mathbf{x})$ for the generating task.
Such time-dependent processes can be summarized as follow
\begin{align}
    \mathbf{x}_t = \alpha_t \bx_* + \sigma_t \eps,
    \label{eq:time_varying}
\end{align}
where $\alpha_t$ is a decreasing function of $t$ and $\sigma_t$ is an increasing function of $t$.
Stochastic interpolants and other flow matching methods \cite{albergo2023building,
albergo2023stochastic, lipman2023flow, liu2022flow} restrict the process~\eqref{eq:time_varying} on $t\in[0,1]$, and set  $\alpha_0=\sigma_1=1$, $\alpha_1=\sigma_0=0$, so that $\bx_t$ interpolates exactly between $\bx_*$ at time $t=0$ and $\eps$ and time $t=1$.
By contrast, score-based diffusion models~\cite{song2021scorebased, kingma2023variational, karras2022elucidating} set both $\alpha_t$ and $\sigma_t$ indirectly through a forward-time stochastic differential equation (SDE) with $\NN(0, \mathbf{I})$ as its equilibrium distribution, i.e. $\bx_t$ converges to $\NN(0, \mathbf{I})$ only if $t \to \infty$. 

Despite the nuances in formulating the stochastic processes $\bx_t$, common to both stochastic interpolants and score-based diffusion models is the observation that $\bx_t$ can be sampled dynamically using either a reverse-time SDE or a probability flow ordinary differential equation (ODE). 

\paragraph{Probability flow ODE.} 
The marginal probability distribution $p_t(\bx)$ of $\bx_t$ in \eqref{eq:time_varying} coincides with the distribution of the probability flow ODE with a velocity field
\begin{equation}
    \label{eq:prob:flow:ode}
    \dot \bX_t = \bv(\bX_t,t),
\end{equation}
where $\bv(\bx,t)$ is given by the conditional expectation
\begin{equation}
    \label{eq:velocity}
    \begin{aligned}
        \bv(\bx,t) &= \E [ \dot \bx_t | \bx_t = \bx],\\
        &= \dot \alpha_t \E [ \bx_* | \bx_t = \bx]+\dot \sigma_t \E [ \eps | \bx_t = \bx].
    \end{aligned}
\end{equation}
The correspondence between $p_t(\bx)$ and~\eqref{eq:prob:flow:ode} and the formulation of~\eqref{eq:velocity} is derived in Appendix~\ref{sec:velocity-form}.
By solving~\eqref{eq:prob:flow:ode} backwards in time from $\bX_{T} = \eps \sim \NN(0, \mathbf{I})$, we can generate samples from $p_0(\bx)$, which approximates the ground-truth data distribution $p(\bx)$. %
We refer to~\eqref{eq:prob:flow:ode} as a \textit{flow-based} generative model.

\paragraph{Reverse-time SDE.} The time-dependent probability distribution $p_t(\bx)$ of $\bx_t$ also coincides with the distribution of the reverse-time SDE~\cite{anderson1979reverse-time}
\begin{equation}
    \label{eq:sde}
    \rd\bX_t = \bv(\bX_t,t) \rd t - \frac12w_t \bs(\bX_t,t) \rd t + \sqrt{w_t} \rd\bar\bW_t,
\end{equation}
where $\bar\bW_t$ is a reverse-time Wiener process, $w_t>0$ is an arbitrary time-dependent diffusion coefficient, $\bv(\bx,t)$ is the velocity defined in \eqref{eq:velocity}, and $\bs(\bx,t)=\nabla \log p_t(\bx)$ is the score. 
Similar to $\bv$, this score is given by the conditional expectation
\begin{equation}
    \label{eq:score}
    \bs(\bx,t) = - \sigma_t^{-1}\E [ \eps | \bx_t = \bx].
\end{equation}
Again, the correspondence between $p_t(\bx)$ and~\eqref{eq:sde} and the formulation of~\eqref{eq:score} is derived in Appendix~\ref{sec:score-form}. 
Solving the reverse SDE~\eqref{eq:sde} backwards in time from $\bX_{T} = \eps \sim \NN(0, \mathbf{I})$ enables generating samples from the approximated data distribution $p_0(\mathbf{x}) \sim p(\bx)$.
We refer to~\eqref{eq:sde} as a \textit{stochastic} generative model.

\paragraph{Design choices.} Score-based diffusion models typically tie the choice of $\alpha_t$, $\sigma_t$, and $w_t$ in~\eqref{eq:sde} to the drift and diffusion coefficients used in the forward SDE that generates $\bx_t$ (see~\eqref{eq:f:sde} below).
The stochastic interpolant framework decouples the formulation of $\bx_t$ from the forward SDE and shows that there is more flexibility in the choices of $\alpha_t$, $\sigma_t$, and $w_t$. 
Below, we will exploit this flexibility to construct generative models that outperform score-based diffusion models on standard benchmarks in image generation task.

\subsection{Estimating the score and the velocity}
\label{sec:score:vel}
Practical use of the probability flow ODE~\eqref{eq:prob:flow:ode} and the reverse-time SDE~\eqref{eq:sde} as generative models relies on our ability to estimate the velocity $\bv(\bx, t)$ and/or score $\bs(\bx, t)$ fields that enter these equations.
The key observation made in score-based diffusion models is that the score can be estimated parametrically as $\bs_\theta(\bx,t)$ using the loss
\begin{equation}
\label{eq:score:loss}
    \LL_{\mathrm{s}}(\theta) = \int_0^T \E[\Vert \sigma_t \bs_\theta(\bx_t, t) + \eps \Vert^2]\rd t.
\end{equation}
This loss can be derived by using~\eqref{eq:score} along with standard properties of the conditional expectation.
Similarly, the velocity in~\eqref{eq:velocity} can be estimated parametrically as $\bv_\theta(\bx,t)$ via the loss
\begin{align}
    \label{eq:velocity-eq-obj}
    \LL_{\mathrm{v}}(\theta) 
    &= \int_0^T \E[\Vert \bv_\theta(\bx_t, t) - \dot\alpha_t \bx_* - \dot\sigma_t \eps\Vert^2] \rd t .
\end{align}
We note that any time-dependent weight can be included under the integrals in both~\eqref{eq:score:loss} and~\eqref{eq:velocity-eq-obj}. 
These weight factors are key in the context of score-based models when $T$ becomes large~\cite{kingma2023understanding}; in contrast, with stochastic interpolants where $T=1$ without any bias, these weights are less important and might impose numerical stability issue (see Appendix~\ref{sec:sbd}). 

\paragraph{Model prediction.} We observed that only one of $\bs_\theta(\bx,t)$ and $\bv_\theta(\bx,t)$ is needed to be estimated in practice. 
This follows directly from the constraint 
\begin{equation}
\label{eq:c}
\begin{aligned}
    \bx & = \E[\bx_t | \bx_t = \bx],\\
    & = \alpha_t \E[\bx_*|\bx_t=\bx] + \sigma_t \E[\eps|\bx_t=\bx],
\end{aligned}
\end{equation}
which can be used to re-express the score~\eqref{eq:score} in terms of the velocity~\eqref{eq:velocity} as
\begin{align}
       \label{eq:v-eps-equivalence}
       \bs(\bx, t) &= \sigma_t^{-1} \frac{\alpha_t \bv(\bx,t) - \dot \alpha_t \bx}{\dot \alpha_t \sigma_t - \alpha_t \dot \sigma_t}.
\end{align}  
We include a detailed derivation in Appendix~\ref{sec:velocity-score-eq}.
Notably, given the simply linear relationship posed by~\eqref{eq:v-eps-equivalence}, we can also express~$\bv(\bx, t)$ in terms of $\bs(\bx, t)$. 
We will use this relation to specify our \textbf{{model prediction}}. 
In our experiments, we typically learn the velocity field $\bv(\bx, t)$ and use it to express the score $\bs(\bx, t)$ when using an SDE for sampling. 

Note that by our definitions $\dot{\alpha}_t < 0$ and $\dot\sigma_t > 0$, so that the denominator of~\eqref{eq:v-eps-equivalence} is never zero.
Yet, $\sigma_t$ vanishes at $t=0$, making the $\sigma_t^{-1}$ in~\eqref{eq:v-eps-equivalence} cause a singularity\footnote{We remark that $\bs(\bx, t)$ can be shown to be non-singular at $t=0$ analytically if the data distribution $p(\bx)$ has a smooth density~\cite{albergo2023stochastic}, though this singularity appears in numerical implementations and losses in general.}.
This suggests the choice $w_t = \sigma_t$ in~\eqref{eq:sde} to cancel this singularity, for which we will explore the performance in the numerical experiments.

\paragraph{Time discretization.} The objective functions specified above are defined over a continuous time domain, as opposed to DDPM which couples the time grid used in learning to that used in sampling. Learning in continuous time allows us to specify a discretization used in sampling \textit{a posteriori}, which allows for flexibility in both sampling efficiency and performance.

\subsection{Specifying the interpolating process}
In~\cref{sec:generative} we present the general definition of interpolants ($\alpha_t$ and $\sigma_t$) for both stochastic interpolant and score-based diffusion. In this section we dive into more details and specify the three choices of interpolants to explore in the experiments.
\label{sec:prob:path}

\begin{figure*}[t]
    \setlength{\parindent}{0.9cm}
     $\xrightarrow[]{\text{ Increasing transformer sizes}}$
    
  \centering
   \includegraphics[width=.45\linewidth]{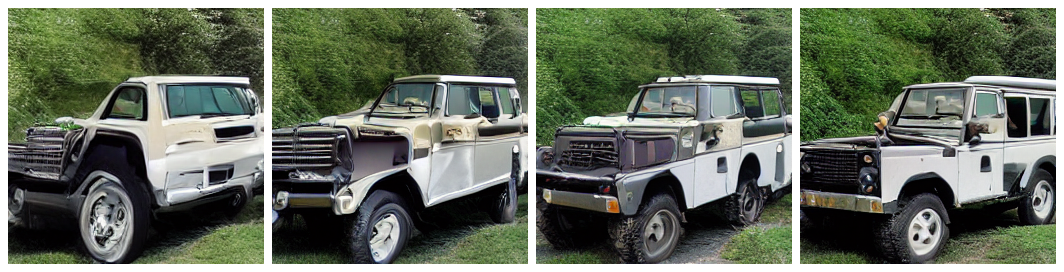}
   \includegraphics[width=.45\linewidth]{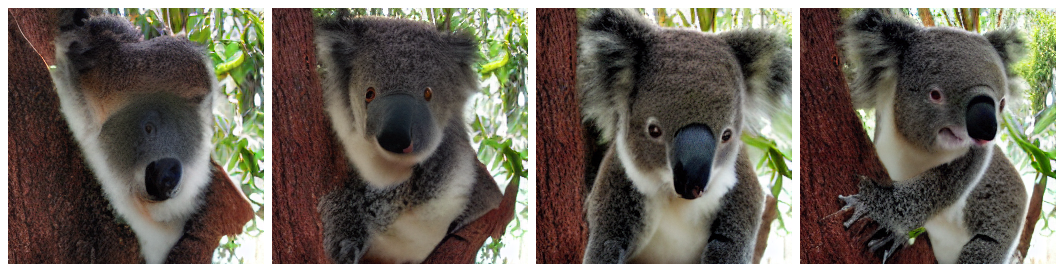}
   \includegraphics[width=.45\linewidth]{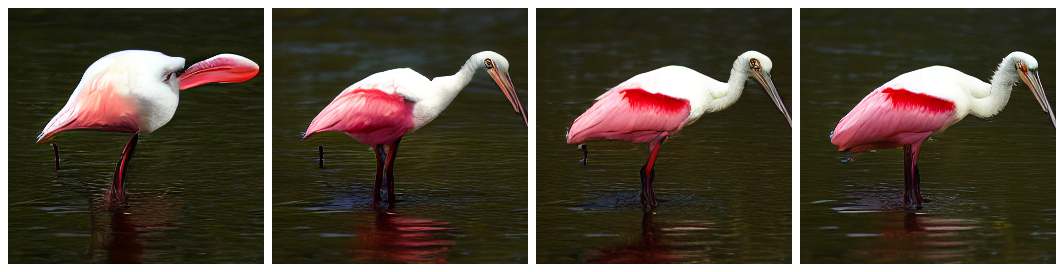}
   \includegraphics[width=.45\linewidth]{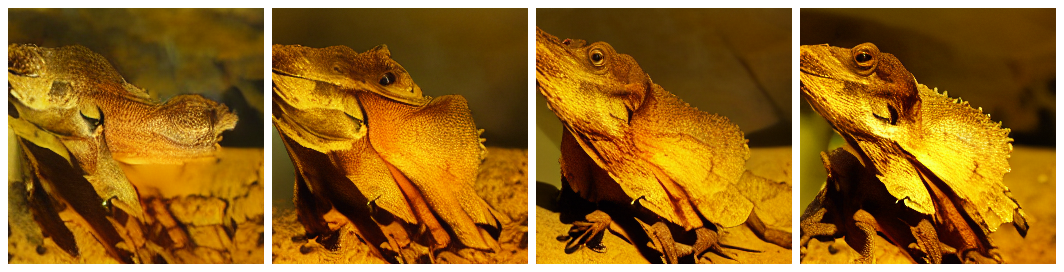}
   \includegraphics[width=.45\linewidth]{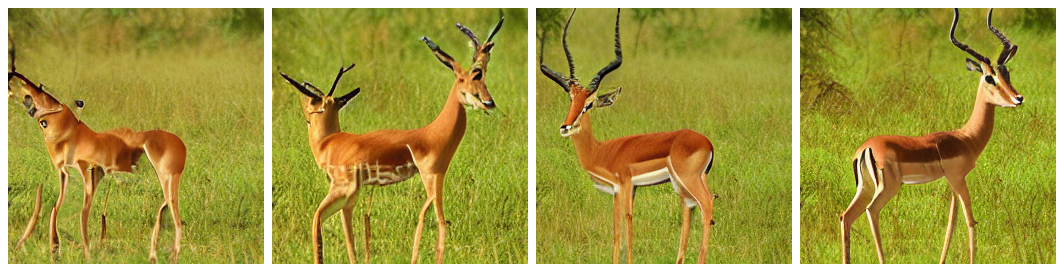}
   \includegraphics[width=.45\linewidth]{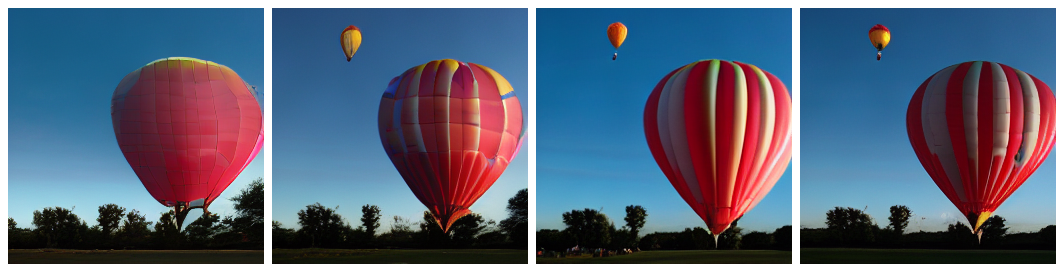}

   \caption{\textbf{Increasing transformer size increases sample quality}. \textit{Best viewed zoomed-in}. We sample from all $4$ of our SiT model (SiT-S, SiT-B, SiT-L and SiT-XL) after 400K training steps using the same latent noise and class label. }
   \label{fig:transformer_scale_samples}
   % \vspace{-1em}
\end{figure*}

\paragraph{Score-based diffusion.} 
We follow~\cite{song2021scorebased} and use the standard variance-preserving (VP) SDE in forward-time
\begin{align}
\label{eq:f:sde}
     \rd\bX_t &= -\frac{1}{2}\beta_t\bX_t \rd t + \sqrt{\beta_t} \rd \bW_t
\end{align}
for some $\beta_t >0$, $\bx_t$'s perturbation kernel $p_t(\bx_t | \bx_0) = \NN(\alpha_t \bx_t, \sigma_t^2 \mathbf{I})$ is defined by 
\begin{align}
    \label{eqn:sbdm_interp}
     \text{SBDM-VP:}\:\: \alpha_t = e^{-\tfrac12 \int_0^t\beta_s \rd s}, \qquad \sigma_t = \sqrt{1- e^{- \int_0^t\beta_s \rd s}}.
\end{align}
The only design flexibility in~\eqref{eqn:sbdm_interp} comes from the choice of $\beta_t$, as 
it determines both
$\alpha_t$ and $\sigma_t$\footnote{
VP is the only linear scalar SDE with an equilibrium distribution~\cite{singhal2023diffuse}; interpolants extend beyond $\alpha_t^2+\sigma_t^2=1$ by foregoing the requirement of an equilibrium distribution.
}.
For example, setting $\beta_t =1$ leads to $\alpha_t = e^{-t}$ and $\sigma_t = \sqrt{1-e^{-2t}}$. 
This choice necessitates taking $T$ sufficiently large~\cite{ho2020denoising} or searching for more appropriate choices of $\beta_t$~\cite{song2021scorebased, chen2023importance, singhal2023diffuse} to reduce the bias. To be specific, such bias comes from the mismatch between the condition $\varepsilon \sim \NN(0, \mathbf{I})$ used in practice for sampling and the density of the process $\bx_1 \not \sim \NN(0, \mathbf{I})$, as stated in~\cref{sec:generative}.

\paragraph{General interpolants.} 
In the stochastic interpolant framework, the process~\eqref{eq:time_varying} is defined explicitly and without any reference to a forward SDE, creating more flexibility in the choice of $\alpha_t$ and $\sigma_t$. 
Specifically, any choice satisfying: 

 \begin{center}
  \begin{tabular}{l}
 (i) $\alpha_t^2 + \sigma_t^2>0$ for all $t\in[0,1]$;\\
  (ii) $\alpha_t$ and $\sigma_t$ are differentiable for all $t\in[0,1]$;\\
  (iii) $\alpha_{1}=\sigma_0 = 0$, $\alpha_{0}=\sigma_1 = 1$;\\
  \end{tabular}
 \end{center}

gives a process that interpolates without bias between $\bx_{t=0} = \bx_*$ and $\bx_{t=1}=\eps$. 
In our numerical experiments, we exploit this design flexibility to test, in particular, the choices
\begin{equation}
   \label{eq:alpha:sig}
   \begin{aligned}
       \text{Linear:} \qquad &\alpha_t = 1-t, \qquad && \sigma_t = t,\\
        \text{GVP:} \qquad &\alpha_t = \cos (\tfrac12\pi t), \qquad && \sigma_t = \sin(\tfrac12\pi t),
   \end{aligned}
\end{equation}
where GVP refers to a generalized VP which has constant variance across time for any endpoint distributions with the same variance.
We note that the fields $\bv(\bx, t)$ and $\bs(\bx, t)$ entering~\eqref{eq:prob:flow:ode} and~\eqref{eq:sde} depend on the choice of $\alpha_t$ and $\sigma_t$, and typically must be specified before learning\footnote{The requirement to
learn and sample under one choice of  path specified by $\alpha_t, \sigma_t$, 
at training time may be relaxed and is explored in~\cite{albergo2023multimarginal}.}.
%.
%
This is in contrast to the diffusion coefficient $w(t)$, as we now describe.

\subsection{Specifying the diffusion coefficient}
\label{sec:diff}

As stated earlier, the SBDM diffusion coefficient used in~\eqref{eq:sde} is usually taken to match that of~\eqref{eq:f:sde}.
That is, one sets $w_t=\beta_t$.
In the stochastic interpolant framework, this choice is again subject to greater flexibility:  \textit{any} $w_t \ge 0$ can be used. 
Interestingly, this choice can be made \textit{after} learning, as it does not affect the velocity $\bv(\bx,t)$ or the score~$\bs(\bx,t)$.
In our experiments, we exploit this flexibility by considering the following choices:

\begin{enumerate}[label=(\roman*)]
    \item $w_t = \sigma_t$; this is used to eliminate the singularity at $t=0$ following the explanation at the end of \cref{sec:score:vel};
    \item $w_t = \sin^2(\pi t)$; this also eliminates the singularity at $t=0$, and allows us to explore the effect of removing diffusivity at times close to $t=1$ in sampling.
    \item $w_t$ can be chosen to minimize an upper bound on the KL divergence $D_{\mathrm{KL}}(p(\bx) \Vert p_0(\bx))$, where $p(\bx)$ denotes the true data distribution and $p_0(\bx)$ refers to the density of $\bx_t$ at $t = 0$. 
    Disregarding the simulation cost of integrating the SDE~\eqref{eq:sde}, it can be shown (see Appendix~\ref{app:wt_kl_bound}) that the following choice of $w_t$ minimizes the KL upper bound: \begin{align}
    \label{eq:w_t:opt}
        w_t = w_t^{\mathrm{KL}} \equiv  2\left(\dot \sigma_t \sigma_t - \frac{\dot \alpha_t \sigma_t^2}{\alpha_t}\right).
    \end{align}
    Under the SBDM-VP interpolant, $w^{\mathrm{KL}}_t$ coincides with $\beta_t$; this aligns with the claim made in~\cite{song2021maximum}.
    \item If the SDE in (iii) becomes hard to integrate because of the magnitude of $w_t^{\mathrm{KL}}$ near $t=1$, one may wish to \textit{regularize} the diffusion coefficient to reduce the integration cost. 
    For example, difficulties may arise for the Linear and GVP interpolants, because $w_t^{\mathrm{KL}} \to \infty$ as $t \to 1$ given the presence of $\alpha_t$ in the denominator of \eqref{eq:w_t:opt}. 
    Including the integration cost of~\eqref{eq:sde}, it can also be shown (see Appendix~\ref{app:wt_kl_bound}) that an optimal regularized $w_t$ is given by
    \begin{align}
    \label{eq:w_t:opt:reg}
        w_t^{\mathrm{KL}, \eta} \equiv w_t^{\mathrm{KL}}\sqrt{\frac{\LL_t}{\LL_t + 2\eta (w_t^{\mathrm{KL}})^2}},
    \end{align}
    where $\LL_t$ is the value of $\LL_{\mathrm{v}}$ in~\cref{sec:score:vel} at time $t$, and $\eta$ is any non-negative constant. 
    With $\eta = 0$, we recover $w_t^{\mathrm{KL}}$. 
    For score models, we first convert to a velocity model following~\eqref{eq:v-eps-equivalence}, then calculate the corresponding $\LL_{\mathrm{v}}$. 
    As $t \to 1$, $ w_t^{\mathrm{KL}, \eta}$ approaches a limit at $ \sqrt{\frac{\LL_{t \to 1}}{2\eta}}$. If $\LL_t$ is defined everywhere on $[0, 1]$, then $w_t^{\mathrm{KL}, \eta}$ will be well-behaved on $[0, 1]$.
\end{enumerate}

\subsection{Interpolant Transformer Architecture}
\label{sec:transformer:arch}
The backbone architecture and capacity of generative models are both crucial for producing high-quality samples. In order to eliminate any confounding factors and focus on our exploration, we strictly follow the standard Diffusion Transformer (DiT)~\cite{peebles2023scalable} and its configurations. This way, we can also test the scalability of our model across various model sizes. 

Here we briefly introduce the model design. Generating high-resolution images with diffusion models can be computationally expensive. Latent diffusion models (LDMs)~\cite{rombach2022high} address this by first downsampling images into a smaller latent embedding space using an encoder $E$, and then training a diffusion model on $z = E(x)$. New images are created by sampling $z$ from the model and decoding it back to images using a decoder $x = D(z)$. 

Similarly, SiT is a latent generative model, and we use the same pre-trained VAE encoder and decoder models originally used in Stable Diffusion~\cite{rombach2022high}. 
SiT processes a spatial input $z$ (shape $32\times32\times4$ for $256\times256\times3$ images) by first `patchifying' it into $T$ linearly embedded tokens of dimension $d$. We always use a patch size of 2 in these models as they achieve the best sample quality. We then apply standard ViT~\cite{dosovitskiy2021image} sinusoidal positional embeddings to these tokens. We use a series of $N$ SiT transformer blocks, each with hidden dimension $d$. 

Our model configurations---SiT-\{S,B,L,XL\}---vary in model size (parameters) and compute (flops), allowing for a model scaling analysis. For class-conditional generation on ImageNet, we use the AdaLN-Zero block~\cite{peebles2023scalable} to process additional conditional information (times and class labels). SiT architectural details are listed in Appendix~\ref{sec:imple}. 

\newcommand{\fancybreak}[0]{
    \noindent~\hfill\noindent\rule{0.9\linewidth}{0.3pt}~\hfill~
}

{
% \vspace{-0.2cm}
\fancybreak
}

The complete SiT design space that we explore consists of the choice of time discretization and the model prediction (\cref{sec:score:vel}), the choice of the interpolant (\cref{sec:prob:path}), the choice of sampler and diffusion coefficient (\cref{sec:diff}), and the model size (\cref{sec:transformer:arch}).

%%%%%%%%%%%%%%%%%%%%%%%%%%%%%%%%%%%%%%%%%%%%%%%%%%%%%%%%%%%%%
%%%%%%%%%%%%%%%%%%%%%%%%% Experiment %%%%%%%%%%%%%%%%%%%%%%%%
%%%%%%%%%%%%%%%%%%%%%%%%%%%%%%%%%%%%%%%%%%%%%%%%%%%%%%%%%%%%%

\section{Experiments}
\label{sec:experiments}
To provide a more detailed answer to the question raised in Table~\ref{tab:FID-50K} and make a fair comparison between DiT and SiT, we gradually transition from a DiT model (discrete, score prediction, VP interpolant) to a SiT model (continuous, velocity prediction, Linear interpolant) and present the impacts on performance. 
\paragraph{Experimental setup.} In the transition experiments, we use SiT-B models trained on $256 \times 256$ image resolution on the ImageNet as our backbone. We fix training steps to be 400K throughout the transition. For solving the ODE~\eqref{eq:prob:flow:ode}, we adopt a fixed step second-order Heun integrator; for solving the SDE~\eqref{eq:sde}, we used a first-order Euler-Maruyama integrator. With both solver choices we limit the number of function evaluations (NFE) to be $250$ to match the number of sampling steps used in DiT. All metrics presented are FID-50K scores evaluated on the ImageNet training set unless otherwise stated. 

We also scale up our SiT model to the XL configuration and train on both $256 \times 256$ and $512 \times 512$ resolution on ImageNet. \textit{We strictly follow the training settings of DiT and did not tune any hyperparameters.}

\subsection{Model Parameterization}
\label{sec:param}

\paragraph{Discrete- to continuous-time.} 

Continuous time training has been previously studied from the perspective of improved likelihood bounds
\cite{song2021scorebased, kingma2023variational}. As mentioned in \Cref{sec:score:vel},  here we focus on the  fact that training in continuous time allows us to decouple discretization choices in sampling from the particular training method, which allows for finding the right discretization for various choices of diffusion coefficients that we are free to choose after training. We observe a marginal performance increase in \Cref{tab:sbdm_ddpm} by switching to continuous time. 

We additionally observe in \Cref{fig:nfe_sampler} that flexibility in integration allows one to trade-off number of functional evaluations and FID performance. 

\paragraph{Model parameterization.}
To clarify the role of the model parameterization in the context of SBDM-VP, we now compare learning (i) a score model using~\eqref{eq:score:loss} ($\mathcal{L}_\mathrm{s}$), (ii)  a weighted score model ($\mathcal{L}_{\mathrm{s}_\lambda}$), or (iii)  a velocity model using~\eqref{eq:velocity-eq-obj}($\mathcal{L}_{\mathrm{v}}$).
We observe a significant performance increase with $\mathcal{L}_{\mathrm{s}_\lambda}$ and $\mathcal{L}_{\mathrm{v}}$ in Table~\ref{tab:model}. 

In accordance with the observation made in~\cite{kingma2023understanding}, we carefully choose a $\lambda(t)$ such that $\lambda_{\mathrm{s}_\lambda}$ is made equivalent to $\lambda_{\mathrm{v}}$. We will provide detailed derivations in Appendix~\ref{sec:score-form}, and demonstrate such $\lambda$ is closely related to the maximum likelihood weighting proposed in~\cite{song2021maximum, vahdat2021scorebased}. Furthermore, we note that $\lambda(t) \to \infty$ as $t \to 0$, thus compensating for the vanishing gradient of the score objective when near the data. This could also account for the performance gain from $\lambda_{\mathrm{s}}$ to $\lambda_{\mathrm{s}_\lambda}$.

\begin{figure}
    % \vspace{-2em}
    \begin{minipage}[t]{.48\linewidth}
        \centering
       \captionof{table}{\textbf{Discrete vs. continuous.}}
        \begin{tabular}{ccccc}
        \toprule
         \textit{ } & \textit{Model} & \textit{Objective} & FID\\
        \midrule
         DDPM & Noise & $\LL_{\mathrm{s}}^N$ & 44.2\\
         SBDM-VP  & Score & $\LL_{\mathrm{s}}$ & \textbf{43.6}\\
         \bottomrule
        \end{tabular}
        \label{tab:sbdm_ddpm}
    \end{minipage}\hfill
    \begin{minipage}[t]{.48\linewidth}
        \centering
        \captionof{table}{\textbf{Effect of parameterizations.}}
        \begin{tabular}{cccc}
        \toprule
        \textit{Interpolant} & \textit{Model} & \textit{Objective} & FID \\
        \midrule
         SBDM-VP &  \text{Score} & $\LL_{\mathrm{s}}$ & 43.6 \\
         SBDM-VP &  \text{Score} & $\LL_{\mathrm{s}_\lambda}$ & \textbf{39.1} \\
         SBDM-VP &  \text{Velocity} & $\LL_{\mathrm{v}}$  & 39.8\\
         \bottomrule
        \end{tabular}
        \label{tab:model}
    \end{minipage}
    % \vspace{-2em}
\end{figure}

\paragraph{Choices of interpolant.}
\cref{sec:sit} highlights that there are many possible ways to build a connection between the data distribution and a Gaussian by varying the choice of $\alpha_t$ and $\sigma_t$ in the definition of the interpolant~\eqref{eq:time_varying}.
To understand the role of this choice, we now study the benefits of moving away from the commonly-used SBDM-VP setup. 
We consider learning a velocity model $\bv(\bx, t)$ with the Linear and GVP interpolants presented in~\eqref{eq:alpha:sig}, which make the interpolation between the Gaussian and the data distribution exact on $[0,1]$. 
We benchmark these models against the SBDM-VP in Table~\ref{tab:interpolant}, where we find that both the GVP and Linear interpolants obtain significantly improved performance.

One possible explanation for this observation is given in~\cref{fig:density_paths_length}, where we see that the path length (transport cost) is reduced when changing from SBDM-VP to GVP or Linear.
We note that this is equivalently reducing curvatures
in the ODE trajectories from SBDM-VP to Linear, which is known to reduce the
time-discretization errors in sampling~\cite{liu2022flow, lee2023curvature}, and thus contributing to the performance.
Numerically, we also note that for SBDM-VP, $\dot \sigma_t = \beta_t e^{-\int_0^t\beta_sds}/(2\sigma_t)$ becomes singular at $t = 0$: this can pose numerical difficulties inside $\LL_{\mathrm{v}}$, leading to difficulty in learning near the data distribution.
This issue does not appear with the GVP and Linear interpolants.

\begin{figure}
    % \vspace{-2em}
    \begin{minipage}[t]{.48\linewidth}
        \centering
       \captionof{table}{\textbf{Effect of interpolant.}}
        \begin{tabular}{cccc}
        \toprule
         \textit{Interpolant} & \textit{Model} & \textit{Objective} & FID \\
        \midrule
         SBDM-VP & \text{Velocity} & $\LL_{\mathrm{v}}$ & 39.8 \\
         Linear &  \text{Velocity} & $\LL_{\mathrm{v}}$ & 34.8  \\
        GVP &  \text{Velocity} & $\LL_{\mathrm{v}}$ & \textbf{34.6} \\
         \bottomrule
        \end{tabular}
        \label{tab:interpolant}
    \end{minipage}\hfill
    \begin{minipage}[t]{.48\linewidth}
        \centering
        \captionof{table}{\textbf{ODE vs. SDE, $w_t = w_t^{\mathrm{KL}}$.}}
        \begin{tabular}{ccccc}
        \toprule
         \textit{Interpolant} & \textit{Model} & \textit{Objective} & ODE & SDE \\
        \midrule
         SBDM-VP & \text{Velocity} & $\LL_{\mathrm{v}}$ &39.8 & 37.8 \\
         Linear &  \text{Velocity} & $\LL_{\mathrm{v}}$ & 34.8 & 33.6 \\
        GVP &  \text{Velocity} & $\LL_{\mathrm{v}}$ & 34.6 & \textbf{32.9}\\
         \bottomrule
        \end{tabular}
        \label{tab:ode_sde}
    \end{minipage}
    % \vspace{-3em}
\end{figure}

\subsection{Deterministic vs stochastic sampling}
\label{sec:samplers}

\begin{figure}
    \begin{minipage}[t]{.48\linewidth}
        \centering
        \includegraphics[width=\linewidth]{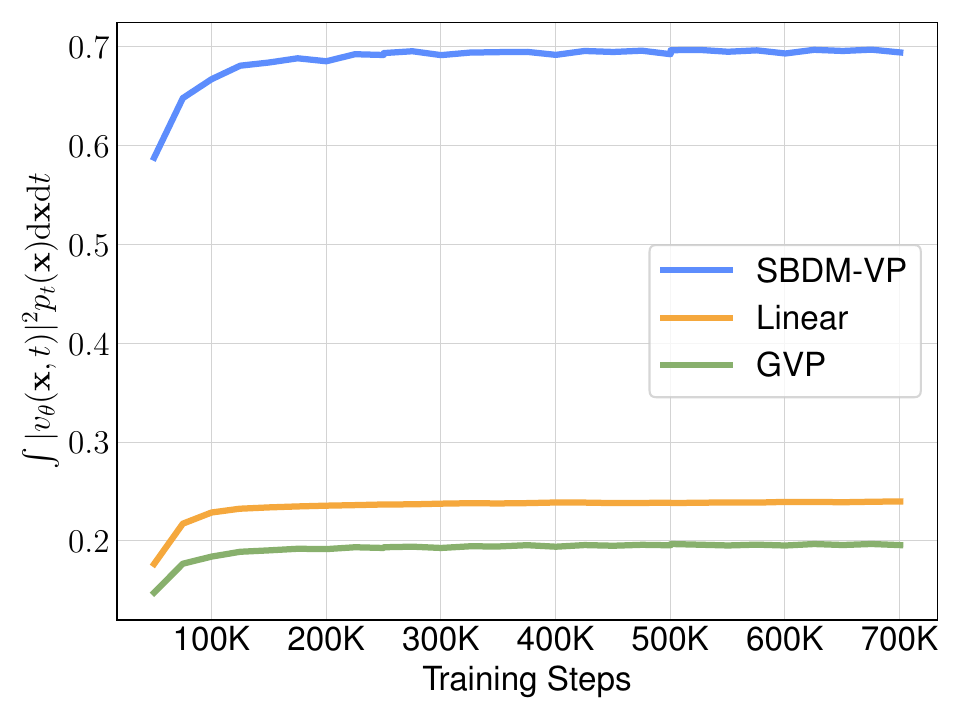}
        \captionof{figure}{\textbf{Path length.} The path length $\mathcal C(v) = \int_0^1 \mathbb E[|\bv(\bx_t,t)|^2] \rd t$ arising from the velocity field at different training steps; each curve is approximated by $10000$ datapoints at each training step.}
       \label{fig:density_paths_length}
    \end{minipage}\hfill
    \begin{minipage}[t]{.48\linewidth}
        \centering
        \includegraphics[width=\linewidth]{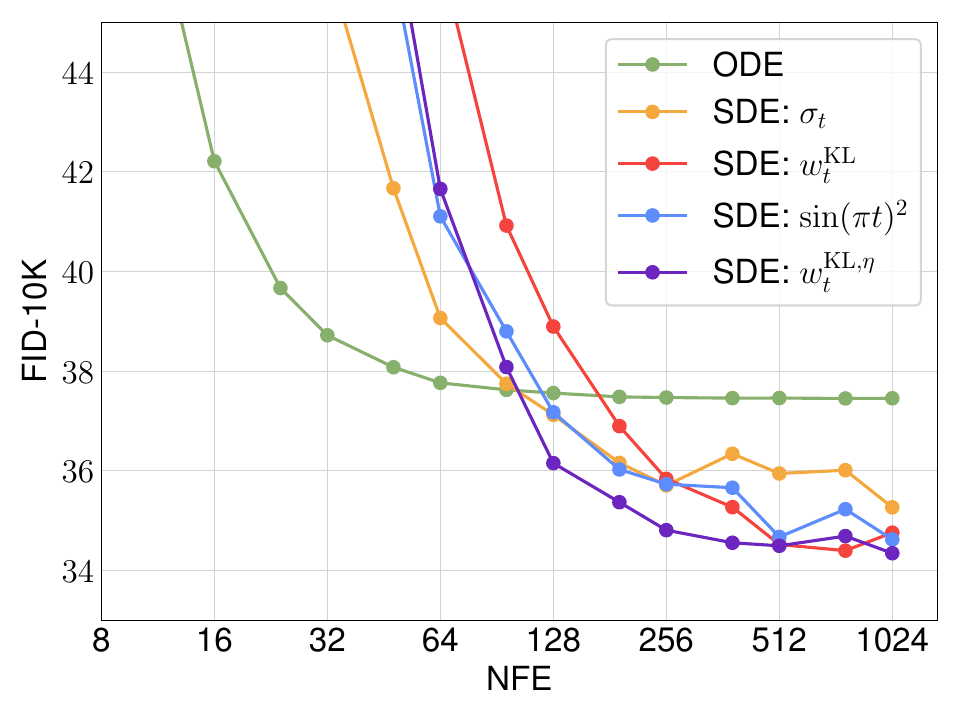}
       \captionof{figure}{\textbf{Comparison of ODE and SDE w/ choices of diffusion coefficients. } We evaluate each sampler using a 400K steps trained SiT-B model with Linear interpolant and learning the $\bv(\bx, t)$.}
       \label{fig:nfe_sampler}
    \end{minipage}
    % \vspace{-2em}
\end{figure}

As shown in~\cref{sec:sit}, given a learned model, we can sample using either the probability flow equation~\eqref{eq:prob:flow:ode} or an SDE~\eqref{eq:sde}.
In \cref{tab:ode_sde}  we illustrate the discrepancy between the two methods when using the same trained velocity model. 
We find performance improvements by sampling with an SDE over the ODE, which is in line with the bounds given in \cite{albergo2023stochastic}: the SDE has better control over the KL divergence between the model density at $t = 0$ and the ground truth data distribution.
We also note that the performance of ODE and SDE integrators may differ under different computation budgets. 
As shown in~\cref{fig:nfe_sampler}, the ODE converges faster with fewer NFE, while the SDE is capable of reaching a much lower final FID score when given a larger computational budget. 

\paragraph{Tunable diffusion coefficient.} Motivated by the improved performance of SDE sampling, we now consider the effect of tuning the diffusion coefficient in inference.
As shown in Table~\ref{tab:general-SDE}, we sweep through all different combinations of our model prediction and interpolant, and present the result. We find that the optimal choice for sampling is both \textit{model prediction} and \textit{interpolant} dependent. 

According to~\cref{sec:diff}, the choice of $w_t = w_t^{\mathrm{KL}}$ would ideally minimize the upper bound for the KL divergence $D_{\mathrm{KL}}(p(\bx) |\Vert p_0(\bx))$ and make the SDE approximate the data distribution more closely, barring integration costs.
This theoretical result is supported by empirical observation for the SBDM-VP and GVP interpolants presented in Table~\ref{tab:general-SDE}. 
For Linear interpolants, the cost-regularized version $w_t^{\mathrm{KL},\eta}$ provides the best FID, because the SDE for the Linear interpolant with $w_t^{\mathrm{KL}}$ becomes hard to integrate at the endpoint.
Generally speaking, the score models perform worse than the velocity models, which may be due to the singularity of the objective in \eqref{eq:score:loss}. Moreover, the efficacy of using $w_t^{\mathrm{KL}}$ in this context is also reduced, for similar reason. For example, reverting~\eqref{eq:v-eps-equivalence} to obtain $\mathrm{v}_\theta(\bx, t)$ from $\mathrm{s}_\theta(\bx, t)$ will result in a singularity at $t = 1$ in $\mathcal L_t$ used to choose $w_t^{\mathrm{KL},\eta}$ in \eqref{eq:w_t:opt:reg}.
Lastly, for SBDM-VP we observe worse result from $w_t^{\mathrm{KL}, \eta}$ as opposed to $w_t^{\mathrm{KL}}$. 
Different from Linear and GVP, as stated in~\cref{sec:diff} and~\cref{sec:param}, $w_t^{\mathrm{KL}}$ is well-defined everywhere on $[0, 1]$ for SBDM-VP, whereas $w_t^{\mathrm{KL}, \eta}$ suffers from the singularity issue posed by $\LL_{\mathrm{v}}$ near $t = 0$.
These observations supports our claim made before, that the optimal choice of $w_t$ will always be \textit{model prediction} and \textit{interpolant} dependent. 

We also note that the influences of different diffusion coefficients can vary across different model sizes. Empirically, we observe the best choice for our SiT-XL is a velocity model with Linear interpolant and sampled with $w_t^{\mathrm{KL}, \eta}$.

\begin{table*}[t!]
    \centering
    \caption{\textbf{Evaluation of our SDE samplers.} The last three columns specify different diffusion coefficients $w_t$. To make the SBDM-VP competitive, we perform evaluation on the weighted score model $\mathcal{L}_{\mathrm{s}_\lambda}$. We mark the optimal $w_t$ for each interpolant.}
    \setlength\tabcolsep{5pt} % Default value: 6pt
    \renewcommand{\arraystretch}{0.85} % Default value: 1
    \begin{tabular}{ccccccc}
    \toprule
   \textit{Interpolant} & \textit{Model} & \textit{Objective} & $w_t = w_t^{\mathrm{KL}} $& $w_t = \sigma_t$ & $w_t = \sin^2(\pi t)$ & $w_t = w_t^{\mathrm{KL}, \eta}$ \\
    \addlinespace
    \toprule
      SBDM-VP & velocity & $\LL_{\mathrm{v}}$ & 37.8 & 38.7 & 39.2 & 41.1\\
     & score & $\LL_{\mathrm{s}_\lambda}$ & \textbf{35.7} &  37.1 & 37.7 & 38.9\\
    \midrule
    GVP & velocity & $\LL_{\mathrm{v}}$ & \textbf{32.9} & 33.4 & 33.6 & 33.2\\
     & score & $\LL_{\mathrm{s}}$ &37.8 & 33.5 & 33.2 & 33.3\\
     \midrule
     Linear &  velocity & $\LL_{\mathrm{v}}$ & 33.6 & 33.5 & 33.3 & \textbf{33.0}\\
     & score &  $\LL_{\mathrm{s}}$ & 41.0 & 35.3 & 34.4 & 34.9\\
    \bottomrule
    \end{tabular}
    \label{tab:general-SDE}
    % \vspace{-1em}
\end{table*}

\subsection{Classifier-free guidance}

Classifier-free guidance (CFG)~\cite{ho2022classifier} often leads to improved performance for score-based models. In this section, we give a concise justification for adopting it on the velocity model, and then empirically show that the drastic gains in performance for DiT case carry across to SiT.

Guidance for a velocity field means that: (i) that the velocity model $\bv_\theta(\bx, t; \by)$ takes class labels $y$ during training, where $y$ is occasionally masked with a null token $\emptyset$; and (ii)  during sampling the velocity used is $\bv_\theta^\zeta(\bx, t; \by) = \zeta \bv_\theta(\bx, t; \by) + (1-\zeta) \bv_\theta(\bx, t; \emptyset)$ for a fixed $\zeta > 0$. 
In Appendix~\ref{sec:guidance}, we show that this indeed corresponds to sampling the tempered density $p(\bx_t)p(\by|\bx_t)^\zeta$ as proposed in~\cite{nichol2021improved}.
Given this observation, one can leverage the usual argument for classifier-free guidance of score-based models.

We observed similar performance improvement with our SiT-XL models under identical computation budget and CFG scale as DiT-X: models.
For SiT-XL $256 \times 256$, we follow identical settings in DiT and train the model for 7M steps. We show samples in~\cref{fig:sota}, and report the result in Table~\ref{tab:FID-50K-IS}. 
For SiT-XL $512 \times 512$, we train the model for 3M steps under the same setting and report the result in Table~\ref{tab:FID-50K-IS}. 
Under both training settings we observe performance advantage of SiT. We display more samples in~\cref{fig:sota} and in Appendix~\ref{sec:add_samples}. 

\begin{table}
  \centering
  \caption{\textbf{Benchmarking class-conditional image generation on ImageNet $256 \times 256$ and $512 \times 512$.} 
  SiT-XL surpasses DiT-XL in both resolutions.}
  \scalebox{0.6}{
  \setlength\tabcolsep{1.5pt}
  \begin{tabular}{lccccc}
    \toprule
    \multicolumn{6}{l}{\textbf{Class-Conditional ImageNet $256 \times 256$}} \\
    \midrule
    Model  & FID$\downarrow$ & sFID$\downarrow$ & IS$\uparrow$ & Precision$\uparrow$ & Recall$\uparrow$ \\
    \midrule
    BigGAN-deep\cite{brock2019large} & 6.95 & 7.36 & 171.4 & \textbf{0.87} & 0.28 \\
    StyleGAN-XL\cite{sauer2022styleganxl} & 2.30 & \textbf{4.02} & 265.12 & 0.78 & 0.53 \\
    \midrule
    Mask-GIT\cite{chang2022maskgit} & 6.18 & - & 182.1 & - & -\\
    \midrule
    ADM\cite{dhariwal2021diffusion} & 10.94 & 6.02 & 100.98 & 0.69 & 0.63 \\
    ADM-G, ADM-U & 3.94 & 6.14 & 215.84 & 0.83 & 0.53 \\
    \arrayrulecolor{black!30}\midrule
    CDM\cite{ho2021cascaded} & 4.88 & - & 158.71 & - & - \\
    \midrule
    RIN\cite{jabri2023scalable} & 3.42 & - & 182.0 & - & - \\
    \midrule 
    Simple Diffusion(U-Net)\cite{hoogeboom2023simple} & 3.76 & - & 171.6 & - & -\\
    Simple Diffusion(U-ViT, L) & 2.77 & - & 211.8 & - & - \\
    \midrule
    VDM++\cite{kingma2023understanding} & 2.12 & - & 267.7 & - & -\\
    \midrule
    DiT-XL(cfg = 1.5)\cite{peebles2023scalable} & 2.27 & 4.60 & 278.24 & 0.83 & 0.57 \\
    \midrule
    \textbf{SiT-XL(cfg = 1.5, ODE)}   & 2.15  & 4.60 & 258.09 & 0.81 & 0.60 \\
    \textbf{SiT-XL(cfg = 1.5, SDE)}   & \textbf{2.06} & 4.49 &  277.50 & 0.83 & 0.59\\
    \arrayrulecolor{black}\bottomrule
  \end{tabular}
  
  \setlength\tabcolsep{1.5pt}
  \begin{tabular}{lccccc}
    \toprule
    \multicolumn{6}{l}{\textbf{Class-Conditional ImageNet $512 \times 512$}} \\
    \midrule
    Model  & FID$\downarrow$ & sFID$\downarrow$ & IS$\uparrow$ & Precision$\uparrow$ & Recall$\uparrow$ \\
    \midrule
    BigGAN-deep\cite{brock2019large} & 8.43 & 8.13 & 177.90 & \textbf{0.88} & 0.29 \\
    StyleGAN-XL\cite{sauer2022styleganxl} & \textbf{2.41} & \textbf{4.06} & 267.75 & 0.77 & 0.52 \\
    \midrule
    Mask-GIT\cite{chang2022maskgit} & 7.32 & - & 156.0 & - & -\\
    \midrule
    ADM\cite{dhariwal2021diffusion} & 23.24 & 10.19 & 58.06 & 0.73 & 0.60 \\
    ADM-G, ADM-U & 3.85 & 5.86 & 221.72 & 0.84 & 0.53 \\
    \arrayrulecolor{black!30}\midrule
    Simple Diffusion(U-Net)\cite{hoogeboom2023simple} & 4.28 & - & 171.0 & - & -\\
    Simple Diffusion(U-ViT, L) & 4.53 & - & 205.3 & - & - \\
    \midrule
    VDM++\cite{kingma2023understanding} & 2.65 & - & 278.1 & - & -\\
    \midrule
    DiT-XL(cfg = 1.5)\cite{peebles2023scalable} & 3.04 & 5.02 & 240.82 & 0.84 & 0.54 \\
    \midrule
    \textbf{SiT-XL(cfg = 1.5, SDE)}   & \textbf{2.62} & 4.18 &  252.21 & 0.84 & 0.57\\
    \arrayrulecolor{black}\bottomrule
  \end{tabular}
  }
  \label{tab:FID-50K-IS}
  % \vspace{-1em}
\end{table}

%%%%%%%%%%%%%%%%%%%%%%%%%%%%%%%%%%%%%%%%%%%%%%%%%%%%%%%%%%%%%
%%%%%%%%%%%%%%%%%%%%%%%% Related Work %%%%%%%%%%%%%%%%%%%%%%%
%%%%%%%%%%%%%%%%%%%%%%%%%%%%%%%%%%%%%%%%%%%%%%%%%%%%%%%%%%%%%

\section{Related Work}
\label{sec:background}

\paragraph{Transformers.} 

The transformer architecture~\cite{vaswani_attention} has emerged as a powerful tool for application domains as diverse as vision~\cite{parmar_image_nodate, dosovitskiy2021image}, language~\cite{zaheer_big_2021, wang_learning_2019}, quantum chemistry~\cite{von_glehn_self-attention_2023}, 
active matter systems~\cite{boffi2023deep},
and biology~\cite{chandra_transformer-based_2023}.
Several works have built on DiT
and have made improvements by modifying the architecture to internally include masked prediction layers \cite{gao2023masked,zheng2023fast}; these choices are orthogonal to this work and may be fruitfully combined in future work.

\paragraph{Training and Sampling in Diffusions.}
Diffusion models arose from \cite{dickstein15, ho2020denoising, song2021scorebased} and have close historical relationship with denoising methods \cite{simoncelli1996, hyvarinen1999, hyvarinen05a}. Various efforts have gone into improving the sampling algorithms behind these methods in the context of DDPM \cite{song2022denoising} and SBDM \cite{song2021maximum, karras2022elucidating}; these are also orthogonal to our studies and may be combined to push for better performance in future work.
Improved Diffusion ODE \cite{zheng2023improved} also studies several combinations of model parameterizations (velocity versus noise)
and paths (VP versus Linear). Unlike our work, they focus on lower dimensional experiments, benchmark with likelihoods, and do not consider SDE sampling.
\paragraph{Interpolants and flow matching.}
Velocity parameterizations using the Linear interpolant were also studied in \cite{lipman2023flow, liu2022flow}, and were generalized to the manifold setting in~\cite{benhamu2022}. 
A trade-off in bounds on the KL divergence between the target distribution and 
the model arises when considering 
sampling with SDEs versus ODE; \cite{albergo2023stochastic} shows that minimizing the objectives presented in this work controls KL for SDEs, but not for ODEs. 
Error bounds for SDE-based sampling with score-based diffusion models are studied in~\cite{sinho_score1, holden_score1, holden_score2, holden_score3}, for ODE-base sampling are explored in \cite{chen2023, benton2023error}, in addition to the Wasserstein bounds provided in \cite{albergo2023building}. 

Other related works make improvements by changing how noise and data are sampled during training.
\cite{tong2023improving, pooladian2023multisample} compute mini-batch optimal couplings between the Gaussian and data distribution to reduce the transport cost and gradient variance;
\cite{albergo2023coupling} instead build the coupling by flowing directly from the conditioning variable to the data for image-conditional tasks.
Finally, various work considers learning a stochastic bridge connecting two arbitrary distributions \cite{peluchetti2022nondenoising, shi2023diffusion, liu2022let, debortoli2021}.
These directions 
are compatible with our investigations; they specify the learning problem for which one can vary the choices of model parameterizations, interpolant schedules, and sampling algorithms.
\paragraph{Diffusion in Latent Space.}
Generative modeling in latent space 
\cite{vahdat2021scorebased,rombach2022high} is a tractable approach for modeling high-dimensional data. The approach has been applied beyond images to video generation \cite{blattmann2023align}, which is a yet-to-be explored and promising application area for velocity trained models. \cite{dao2023flow} also train velocity models in the latent space of the pre-trained Stable Diffusion VAE. They demonstrate promising results for the DiT-B backbone with a final FID-50K of 4.46; their study was one motivation for
the investigation in this work regarding which aspects of these models contribute to the gains in performance over DiT.

%%%%%%%%%%%%%%%%%%%%%%%%%%%%%%%%%%%%%%%%%%%%%%%%%%%%%%%%%%%%%
%%%%%%%%%%%%%%%%%%%%%%%%% Conclusion %%%%%%%%%%%%%%%%%%%%%%%%
%%%%%%%%%%%%%%%%%%%%%%%%%%%%%%%%%%%%%%%%%%%%%%%%%%%%%%%%%%%%%

\section{Conclusion}
In this work, we have presented Scalable Interpolant Transformers, a simple and powerful framework for image generation tasks. 
Within the framework, we explored the tradeoffs between a number of key design choices: the choice of a continuous or discrete-time model, the choice of interpolant, the choice of model prediction, and the choice of diffusion coefficient.
We highlighted the advantages and disadvantages of each choice and demonstrated how careful decisions can lead to significant performance improvements.
Many concurrent works~\cite{meng2022sdedit, gupta2023photorealistic, liu2023zero1to3, jakab2023farm3d} explore similar approaches in a wide variety of downstream tasks, and we leave the application of SiT to these tasks for future works.

\clearpage

% ---- Acknowledgement ----
\section*{Acknowledgements} We would like to thank Adithya Iyer, Sai Charitha Akula, Fred Lu, Jiatao Gu, and Edwin P. Gerber for helpful discussions and feedback. The research is partly supported by the Google TRC program.

% ---- Bibliography ----
%
% BibTeX users should specify bibliography style 'splncs04'.
% References will then be sorted and formatted in the correct style.
%
\bibliographystyle{splncs04}
\bibliography{main}

\clearpage 
\clearpage
\setcounter{page}{1}
\onecolumn
\appendix
% \title{Supplement}
% \maketitle
% \authorrunning{N.Ma et al.}
\section{Proofs}

In all proofs below, we use $\cdot$ for dot product and assume all bold notations ($\bx$, $\eps$, etc.) are real-valued vectors in $\mathbb{R}^d$. Most proofs are derived from 
\cite{albergo2023stochastic}.

\subsection{Proof of the probability flow ODE~\eqref{eq:prob:flow:ode} with the velocity in~\cref{eq:velocity}. }
\label{sec:velocity-form}

Consider the time-dependent probability  density function (PDF) $p_t(\bx)$ of $\bx_t= \alpha_t \bx_* + \sigma_t \eps$ defined in~\cref{eq:time_varying}. By definition, its  characteristic function $\hat{p}_t(\bk) = \int_{\mathbb{R}^d} e^{i\bk \cdot \bx} p_t(\bx)\rd \bx$ is given by
\begin{align}
    \hat{p}_t(\bk) = \E[e^{i\bk \cdot \bx_t}]
    \label{eq:pt-ch-func}    
\end{align}
where $\E$ denotes expectation over $\bx_*$ and $\eps$.
Taking time derivative on both sides, 
and using the tower property of conditional expectation, we have  
\begin{align}
    \partial_t \hat{p}_t(\bk) 
    &= i\bk \cdot \E[\dot\bx_t e^{i\bk \cdot \bx_t}]\\
     &= i\bk \cdot \E_{\bx\sim p_t}[\E[\dot\bx_t e^{i\bk \cdot \bx_t} | \bx_t = \bx]] \\
    &= i\bk \cdot \E_{\bx\sim p_t}[\E[(\dot \alpha_t \bx_* + \dot \sigma_t \eps) e^{i\bk \cdot \bx_t} | \bx_t = \bx]] \\
    &= i\bk \cdot \E_{\bx\sim p_t}[\E[(\dot \alpha_t \bx_* + \dot \sigma_t \eps) | \bx_t = \bx] e^{i\bk \cdot \bx}] \\
    &= i\bk \cdot \E_{\bx\sim p_t}[\bv(\bx, t) e^{i\bk \cdot \bx}]
    \label{eq:de-pt-ch-func}
\end{align}
where $\bv(\bx, t) = \E[(\dot \alpha_t \bx_* + \dot \sigma_t \eps) | \bx_t = \bx]= \dot \alpha_t\E[ \bx_* | \bx_t = \bx] + \dot \sigma_t \E[\eps | \bx_t=\bx]$ is the velocity defined in~\cref{eq:velocity}.
Explicitly, \cref{eq:de-pt-ch-func} reads
\begin{align}
    \partial_t \int_{\mathbb{R}^d} e^{i\bk \cdot \bx} p_t(\bx)\rd \bx 
    = i\bk \cdot \int_{\mathbb{R}^d} \bv(\bx, t) e^{i\bk \cdot \bx} p_t(\bx)\rd \bx 
    \label{eq:de-pt-ch-func:1}
\end{align}
from which we deduce
\begin{align}
    \int_{\mathbb{R}^d} e^{i\bk \cdot \bx} \partial_t p_t(\bx)\rd \bx 
    &=  \int_{\mathbb{R}^d}  \bv(\bx, t) \cdot \nabla_{\bx} [e^{i\bk \cdot \bx}] p_t(\bx)\rd \bx \\
    &= -\int_{\mathbb{R}^d} \nabla_{\bx} \cdot[\bv(\bx, t)p_t(\bx)] e^{i\bk \cdot \bx} \rd \bx
    \label{eq:de-pt-ch-func:2}
\end{align}
where  $\nabla_\bx \cdot [\bv p_t] = \sum_{i=1}^d \frac{\partial }{\partial x_i} [v_i p_t]$ is the divergence operator and we used integration by parts to get the second equality. 
By the properties of Fourier transform, \cref{eq:de-pt-ch-func:2} implies that $p_t(\bx)$ satisfies the transport equation
\begin{align}
    \partial_t p_t(\bx) + \nabla_\bx \cdot (\bv(\bx,t) p_t(\bx)) = 0.
    \label{eq:transport-eq}
\end{align}
Solving this equation by the method of characteristic leads to probability flow ODE~\eqref{eq:prob:flow:ode}. 

\subsection{Proof of the SDE~\eqref{eq:sde}}
\label{sec:SDE-marginal}
We show that the SDE \eqref{eq:sde} has marginal density $p_t(\bx)$ with any choice of $w_t \ge 0$.  To this end, recall that solution to the SDE $$
    d\bX_t = [\bv(\bX_t,t) - \frac12w_t \bs(\bX_t,t)] dt + \sqrt{w_t} d\bar\bW_t
$$
has a PDF that satisfies the Fokker-Planck equation \begin{align}
    \partial_t p_t(\bx) =
    -\nabla_\bx& \cdot \big([\bv(\bx,t) - \frac12w_t \bs(\bx,t)] p_t(\bx)\big) - \frac{1}{2}w_t\Delta_\bx  p_t(\bx)
    \label{eq:fokker-planck}
\end{align}
where $\Delta_{\bx}$ is the Laplace operator defined as $\Delta_\bx   =  \nabla_\bx \cdot \nabla_\bx= \sum_{i=0}^d \frac{\partial^2}{\partial x_i^2}$. 
Reorganizing the equation and usng the definition of the score $\bs(\bx,t) = \nabla_{\bx} \log p_t(\bx) = p_t^{-1}(\bx) \nabla_{\bx} p_t(\bx)$, we have \begin{align}
    \partial_t p_t(\bx) &= 
    \underbrace{
        -\nabla_\bx \cdot [\bv(\bx,t)p_t(\bx)]
    }_{\text{$=\partial_t p_t(\bx)$ by \cref{eq:transport-eq}}}
    + \frac12w_t  \nabla_\bx \cdot [
        \underbrace{
            \nabla_\bx \log p_t(\bx)  p_t(\bx)
        }_{\text{$=\nabla_\bx p_t(\bx)$}}] 
    - \frac{1}{2}w_t \Delta_\bx  p_t(\bx) \\
 \implies \qquad   0 &= \frac{1}{2}w_t\nabla_\bx \cdot \nabla_\bx p_t(\bx) - \frac{1}{2}w_t \Delta_\bx p_t(\bx)
\end{align}
By definition of Laplace operator, the last equation holds for any $w_t \ge 0$. When $w_t = 0$, the Fokker-Planck equation reduces to a continuity equation, and the SDE reduces to an ODE, so the connection trivially holds.

\subsection{Proof of the expression for the score in \cref{eq:score}}
\label{sec:score-form}
We show that $\bs(\bx, t) = -\sigma_t^{-1}\E[\eps | \bx_t = \bx]$. Letting $\hat{f}(\bk, t) = \E[\eps e^{i\sigma_t \bk \cdot \eps}]$, we have \begin{align}
    \hat{f}(\bk, t) = -\frac{i}{\sigma_t} \nabla_\bk \E[e^{i\sigma_t \bk \cdot \eps}]
\end{align}
Since $\eps \sim \NN(0, \mathbf{I})$, we can compute the expectation explicitly to obtain
\begin{align}
    \hat{f}(\bk, t) &= -\frac{i}{\sigma_t} (\nabla_\bk e^{-\frac{1}{2}\sigma_t^2|\bk|^2}) \\
    &= i\sigma_t \bk e^{-\frac{1}{2}\sigma_t^2 |\bk|^2}
\end{align}
Since $\bx_*$ and $\eps$ are independent random variable, we have \begin{align}
    \E[\eps e^{i\bk \cdot \bx_t}] &= \hat{f}(\bk, t)\E[e^{i\alpha_t \bk \cdot \bx_*}]
    = i\sigma_t \bk \underbrace{e^{-\frac{1}{2}\sigma_t^2 |\bk|^2}\E[e^{i\alpha_t \bk \cdot \bx_*}]}_{\text{combine this}}
    = i\sigma_t \bk \hat{p}_t(\bk)
    \label{eq:eps-exp-pt}
\end{align}
where $\hat{p}_t(\bk)$ is the characteristic function of $\bx_t= \alpha_t \bx_* + \sigma_t \eps$ defined in \cref{eq:pt-ch-func}. 
The left hand-side of this equation can also be written as: 
\begin{align}
    \E[\eps e^{i\bk \cdot \bx_t}] &= \int_{\R^d} \E[\eps e^{i\bk \cdot \bx_t} | \bx_t = \bx]p_t(\bx) \rd \bx \\
    &= \int_{\R^d} \E[\eps | \bx_t = \bx] e^{i\bk \cdot \bx} p_t(\bx) \rd \bx,
    \label{eq:eps-exp-cond}
\end{align}
whereas the right hand-side is
\begin{align}
    i\sigma_t \bk \hat{p}_t(\bk) &= i\sigma_t \bk \int_{\R^d}  e^{i\bk \cdot \bx} p_t(\bx) \rd \bx\\
    & = \sigma_t  \int_{\R^d}  \nabla_{\bx} [e^{i\bk \cdot \bx}] p_t(\bx) \rd \bx\\
    & = - \sigma_t  \int_{\R^d}   e^{i\bk \cdot \bx} \nabla_{\bx} p_t(\bx) \rd \bx\\
    & = - \sigma_t  \int_{\R^d}   e^{i\bk \cdot \bx} \bs(\bx,t) p_t(\bx) \rd \bx
    \label{eq:eps-exp-cond:aa}
\end{align}
where we used integration by parts to get the third equality, and again the definition of the score to get the last.

Comparing~\cref{eq:eps-exp-cond} and \cref{eq:eps-exp-cond:aa} we deduce that, when $\sigma_t \not=0$,
\begin{align}
    \bs(\bx, t) = \nabla_\bx \log p_t(\bx) &= -\sigma_t^{-1} \E[\eps | \bx_t = \bx]
\end{align}
Further, setting $w_t$ to $\sigma_t$ in~\cref{eq:sde} gives \begin{align}
    \frac{1}{2}w_t \bs(\bx_t, t) = -\frac{1}{2}\E[\eps | \bx_t = \bx]
\end{align}
for all $t\in [0,1]$. This bypass the constraint of $\sigma_t \neq 0$ and effectively eliminate the singularity at $t=0$.

\subsection{Proof of \cref{eq:v-eps-equivalence}}
\label{sec:velocity-score-eq}
We note that there exists a straightforward connection between $\bv(\bx, t)$ and $\bs(\bx, t)$. From \cref{eq:time_varying}, we have \begin{align}
    \bv(\bx, t) &= \dot\alpha_t \E[\bx_* | \bx_t = \bx] + \dot \sigma_t \E[\eps | \bx_t = \bx] \\
    &= \dot\alpha_t \E[\frac{\bx_t - \sigma_t \eps}{\alpha_t} | \bx_t = \bx] + \dot \sigma_t \E[\eps | \bx_t = \bx] \\
    &= \frac{\dot \alpha_t}{\alpha_t} \bx + (\dot \sigma_t - \frac{\dot \alpha_t \sigma_t}{\alpha_t})\E[\eps | \bx_t = \bx] \\
    &= \frac{\dot \alpha_t}{\alpha_t} \bx + (\dot \sigma_t - \frac{\dot \alpha_t \sigma_t}{\alpha_t})(-\sigma_t \bs(\bx, t)) \\
    &= \frac{\dot \alpha_t}{\alpha_t} \bx - \lambda_t\sigma_t \bs(\bx, t)
    \label{eq:vel:score}
\end{align}
where we defined 
\begin{equation}
    \label{eq:lambda}
    \lambda_t = \dot \sigma_t - \frac{\dot \alpha_t \sigma_t}{\alpha_t}.
\end{equation}
Given \cref{eq:vel:score} is linear in terms of $\bs$, reverting it will lead to~\cref{eq:v-eps-equivalence}.

Note that we can also plug \cref{eq:vel:score} into the loss $\LL_{\mathrm{\bv}}$ in \cref{eq:velocity-eq-obj}  to deduce that
\begin{align}
    \LL_{\mathrm{v}}(\theta) &= \int_0^T \E[\Vert \underbrace{\frac{\dot \alpha_t}{\alpha_t} \bx}_{\mathclap{\text{Expand to $\bx_t = \alpha_t\bx_* + \sigma_t \eps$}}} + \lambda_t(-\sigma_t\bs_\theta(\bx_t, t)) - \dot\alpha_t \bx_* - \dot\sigma_t \eps\Vert^2] \rd t \\
    &= \int_0^T \E[\Vert \dot \alpha_t \bx_* + \frac{\dot \alpha_t \sigma_t}{\alpha_t}\eps + \lambda_t(-\sigma_t\bs_\theta(\bx_t, t)) - \dot\alpha_t \bx_* - \dot\sigma_t \eps\Vert^2] \rd t \\
    &= \int_0^T \E[\Vert  \lambda_t(-\sigma_t\bs_\theta(\bx_t, t)) - \lambda_t\eps\Vert^2] \rd t \\
    &= \int_0^T \lambda_t^2\E[\Vert  \sigma_t\bs_\theta(\bx_t, t) + \eps\Vert^2] \rd t \\
    &\equiv \LL_{\mathrm{s}_\lambda}(\theta)
\end{align}
which defines the weighted score objective $\LL_{\mathrm{s}_\lambda}(\theta)$. This observation is consistent with the claim made in \cite{kingma2023understanding} that the score objective with different monotonic weighting functions coincides with losses for different model parameterizations. In \cref{sec:sbd} we show that $\lambda_t$ corresponds to the square of the maximum likelihood weighting proposed in \cite{song2021maximum} and \cite{vahdat2021scorebased}.

\subsection{Proof for the optimal $w_t$ for tightening the KL bound \label{app:wt_kl_bound} }
Lemma 2.22 in~\cite{albergo2023stochastic} asserts that: 
\begin{align}
    \label{eq:SDE-KL}
    D_{\mathrm{KL}}(p(\bx) \Vert p_\theta(\bx)) \leq \frac{1}{2}\int_0^1 w_t^{-1}\int_{\Omega} |b(\bx, t) - b_\theta(\bx, t)|^2 p_t(\bx) \rd t \rd \bx
\end{align}
where $p(\bx)$ denotes the true data distribution, $p_\theta(\bx)$ denotes the approximated data distribution by our model at time $t = 0$, and $p_t$ corresponds to the marginal density in~\cref{sec:SDE-marginal}. 
We further use $b$ and $b_\theta$ to refer to the ground truth and approximated drift for the reverse SDE, respectively; that is, $b(\bx, t) = v(\bx, t) - \frac{1}{2}w_ts(\bx, t)$. 
Following~\cref{sec:velocity-score-eq}, $b(\bx, t)$ can be expressed in terms of $v(\bx, t)$ \begin{align}
    \label{eq:SDE-drift}
    b(\bx, t) = (1 + \frac{1}{2}\frac{w_t}{\lambda_t\sigma_t})v(\bx, t) - \frac{1}{2}\frac{w_t\dot\alpha_t}{\alpha_t\lambda_t\sigma_t}x
\end{align}
and similarly for $b_\theta(\bx, t)$. Plug back, \cref{eq:SDE-KL} becomes \begin{align}
    \label{eq:SDE-KL-velocity}
    D_{\mathrm{KL}}(p(\bx) \Vert p_\theta(\bx)) \leq \frac{1}{2}\int_0^1 w_t^{-1}(1 + \frac{1}{2}\frac{w_t}{\lambda_t\sigma_t})^2\int_{\Omega} |v(\bx, t) - v_\theta(\bx, t)|^2 p_t(\bx) \rd t \rd \bx
\end{align}
Since $v(\bx, t) = \E[\dot \bx | \bx_t = \bx]$ from~\cref{eq:time_varying}, we have \begin{align}
    \label{eq:velocity-loss}
    \int_{\Omega} |v(\bx, t) - v_\theta(\bx, t)|^2 p_t(\bx) \rd \bx &= \E[|v(\bx, t) - v_\theta(\bx, t)|^2] \nonumber\\
    &\leq \E[|\dot \bx - v_\theta(\bx, t)|^2] \equiv \LL_t\
\end{align}
where $\LL_t$ is the loss at time $t$ of our model after optimization. With~\cref{eq:SDE-KL-loss} we can further simplify~\cref{eq:velocity-loss} to be \begin{align}
    \label{eq:SDE-KL-loss}
    D_{\mathrm{KL}}(p(\bx) \Vert p_\theta(\bx)) \leq \frac{1}{2}\int_0^1 w_t^{-1}(1 + \frac{1}{2}\frac{w_t}{\lambda_t\sigma_t})^2 \LL_t \rd t
\end{align}
We note the minimum of the integrand in~\cref{eq:SDE-KL-loss} is achieved at $w_t = 2\lambda_t\sigma_t$ with a value of $2\frac{\LL_t}{\lambda_t\sigma_t}$. We note that such $w_t$ is the exact choice of $w_t^{\mathrm{KL}}$ in~\cref{sec:diff}. 

For SBDM-VP interpolant, such $w_t^{\mathrm{KL}}$ coincides with $\beta_t$ in~\cite{song2021scorebased}, and is well defined and positive everywhere on $[0, 1]$. For GVP and Linear interpolant however, this diffusion coefficient is zero at $t = 0$ and infinity at $t = 1$. 
Since $\sigma_t = O(t)$ at $t = 0$, the integrand $2\frac{\LL_t}{\lambda_t\sigma_t}$ is not integrable at $t = 0$, making the bound in~\cref{eq:SDE-KL-loss} trivially $\infty$ unless $\lim_{t\to 0} \LL_t = 0$. 

We note the bound proposed in~\cref{eq:SDE-KL-loss} does not account for the cost of time-integration of the SDE. Assuming that the non-uniform integration step  $\Delta t$ one must take to maintain a given precision is inversely proportional to $w_t$, that is $\Delta t = O(w_t^{-1})$, we can account for the integration cost by adding a term to~\cref{eq:SDE-KL-loss} 
\begin{align}
    \label{eq:SDE-KL-time-error}
    D_{\mathrm{KL}}(p(\bx) \Vert p_\theta(\bx)) \leq \frac{1}{2}\int_0^1 w_t^{-1}\left(1 + \frac{1}{2}\frac{w_t}{\lambda_t\sigma_t}\right)^2 \LL_t \rd t + \eta \int_0^1 w_t \rd t
\end{align}
where $\eta > 0$ is a parameter that controls the integration error: the higher the $\eta$, the smaller the cost but the higher the error, and vice-versa. The minimum of the integrand in~\cref{eq:SDE-KL-time-error} is \begin{align}
    \min_{w_t} \left(  w_t^{-1}\left(1 + \frac{1}{2}\frac{w_t}{\lambda_t \sigma_t}\right)^2\LL_t + \eta w_t\right) = \frac{2\LL_t}{\lambda_t \sigma_t}\left(1 + \sqrt{\frac{4\eta\lambda_t^2\sigma_t^2 + \LL_t}{\LL_t}}\right)
\end{align}
and it is achieved at \begin{align}
    w_t = 2\lambda_t \sigma_t \sqrt{\frac{\LL_t}{4\eta \lambda_t^2\sigma_t^2 + \LL_t}}
\end{align}
This is exactly the choice of $w_t^{\mathrm{KL}, \eta}$ in~\cref{sec:diff}. We note that such diffusion coefficient is well defined everywhere on $[0,1]$ if $\LL_t$ is also well defined everywhere, and as $t \to 1$, it approaches a finite limit at $\sqrt{\frac{\LL_{t\to1}}{2\eta}}$. 

We also note that the integrand in~\cref{eq:SDE-KL-time-error} would still be $\infty$ at time $0$ given the $\frac{1}{\lambda_t\sigma_t}$, unless $\lim_{t\to 0}\LL_t = 0$. 
Theoretically, this is not an unreasonable assumption for both coefficients, as we know the closed form of $v(\bx, 0) = \E[\dot \bx_0 | \bx_{t = 0} = \bx] = \dot \alpha_0 \E[\bx_*]$ and could optimize our model $v_\theta(\bx, t)$ to directly approximate this value at $t = 0$. 
In practice, we found the numerical stability of $w_t^{\mathrm{KL}, \eta}$ could lead to better results. 

\section{Connection with Score-based Diffusion}
\label{sec:sbd}
As shown in \cite{song2021scorebased}, the reverse-time SDE from \cref{eq:f:sde} is \begin{align}
    \rd \bX_t &= [-\frac{1}{2}\beta_t\bX_t - \beta_t\bs(\bX_t, t)] \rd t + \sqrt{\beta_t} \rd \bar{\bW}_t
    \label{eq:vp-rev-sde}
\end{align}  
Let us show this SDE is \cref{eq:sde} for the specific choice $w_t = \beta_t$. To this end, notice that the solution $\bX_t$ to \cref{eq:vp-rev-sde} for the initial condition $\bX_{t=0} = \bx_*$ with $\bx_*$ fixed is Gaussian distributed with mean and variance given respectively by \begin{align}
    \E [\bX_t ]& = e^{-\tfrac12 \int_0^t\beta_s \rd s} \bx_*\equiv  \alpha_t \bx_* \\
    \text{var} [\bX_t] &  = 1- e^{- \int_0^t\beta_s \rd s} \equiv \sigma^2_t
    \label{eq:sde-mean-var}
\end{align}
Using \cref{eq:vel:score}, the velocity of the score-based diffusion model can therefore be expressed as \begin{align}
    \bv(\bx, t) &= -\frac{1}{2}\beta_t \bx + (-\frac{1}{2}\beta_t(1 - e^{- \int_0^t\beta_s \rd s}) - \frac{1}{2}\beta_t e^{- \int_0^t\beta_s \rd s})\bs(\bx, t) \\
    &= -\frac{1}{2}\beta_t \bx - \frac{1}{2}\beta_t \bs(\bx, t)
    \label{eq:vp-velocity}
\end{align}
we see that $2\lambda_t\sigma_t$ is precisely $\beta_t$, making $\lambda_t$ correspond to the square of maximum likelihood weighting proposed in \cite{song2021scorebased}.
Further, if we plug \cref{eq:vp-velocity} into \cref{eq:sde}, we arrive at \cref{eq:vp-rev-sde}.

\paragraph{A useful observation for choosing velocity versus noise model.} We see that in the velocity model, all of the path-dependent terms ($\alpha_t$, $\sigma_t$) are inside the squared loss, and in the score model, the terms are pulled out (apart from the necessary $\sigma_t$ in score matching loss) and get squared due to coming out of the norm. So which is more stable depends on the interpolant. In the paper we see that for SBDM-VP, due to the blowing up behavior of $\dot \sigma_t$ near $t = 0$, both $\LL_{\mathrm{v}}$ and $\LL_{\mathrm{s}_\lambda}$ are unstable. 

Yet, shown in \cref{tab:model}, we observed better performance with $\LL_{\mathrm{s}_\lambda}$ for SBDM-VP, as the blowing up $\lambda_t$ near $t = 0$ will compensate for the diminishing gradient inside the squared norm, where $\LL_{\mathrm{v}}$ would simply experience gradient explosion resulted from $\dot \sigma_t$. 
The behavior is different for the Linear and GVP interpolant, where the source of instability is $\alpha^{-1}_t$ near $t = 1$. We note $\LL_{\mathrm{v}}$ is stable since $\alpha^{-1}_t$ gets cancelled out inside the squared norm, while in $\LL_{\mathrm{s}_\lambda}$ it remains in $\lambda_t$ outside the norm.

\section{Sampling with Guidance}
\label{sec:guidance}

Let $p_t(\bx|\by)$ be the density of $\bx_t= \alpha_t \bx_* + \sigma_t \eps$ conditioned on some extra variable $\by$. By argument similar to the one given in \cref{sec:velocity-form}, it is easy to see that $p_t(\bx|\by)$ satisfies the transport equation (compare \cref{eq:transport-eq})
\begin{equation}
    \label{eq:transport:c}
    \partial_t p_t (\bx|\by)+ \nabla_\bx \cdot (\bv(\bx,t|\by) p_t(\bx,|\by)) = 0,
\end{equation}
where (compare \cref{eq:velocity})
\begin{equation}
\label{eq:velocity:c}
 \bv(\bx,t|\by) = \E [ \dot \bx_t | \bx_t = \bx, \by] = \dot \alpha_t \E [ \bx_*|\bx_t = \bx, \by] + \dot \sigma_t \E[\eps|\bx_t = \bx, \by]
\end{equation}
Proceeding as in \cref{sec:score-form} and \cref{sec:velocity-score-eq}, it is also easy to see that the score $\bs(\bx,t|\by) = \nabla_{\bx}\log p_t(\bx|\by)$ is given by (compare \cref{eq:score})
\begin{equation}
    \label{eq:score:c}
    \bs(\bx,t|\by) = - \sigma_t^{-1} \E [\eps|\bx_t = \bx,\by ]
\end{equation}
and that $\bv(\bx,t|\by)$ and $\bs(\bx,t|\by)$ are related via (compare \cref{eq:vel:score})
\begin{align}
    \bv(\bx, t|\by) = \frac{\dot \alpha_t}{\alpha_t} \bx - \lambda_t\sigma_t \bs(\bx, t|\by)
    \label{eq:vel:score:c}
\end{align}
% and $\bv(\bx,t)$ and $\bs(\bx, t)$ be the formulation derived in \cref{sec:velocity-form} and \cref{sec:score-form} respectively. 
% We can show that we can express $\bv$ in terms of
% the $\bs$, along with $\alpha,\sigma$:
% \begin{align}
%     v(x,t) 
%     &=
% \frac{\dot \alpha_t}{\alpha_t} x
% +
% \Big(
% \dot \sigma_t
% -
% \frac{\dot \alpha_t}{\alpha_t}
% \sigma_t
% \Big)
% \E[\varepsilon | \bx_t=x] \\
%     &=
% \frac{\dot \alpha_t}{\alpha_t} x
% +
% \Big(
% \dot \sigma_t
% -
% \frac{\dot \alpha_t}{\alpha_t}
% \sigma_t
% \Big)
% (-\bsx,t) \sigma_t)\\
%     &=
% \frac{\dot \alpha_t}{\alpha_t} x
% +
% \Big(
% \frac{\dot \alpha_t}{\alpha_t}
% \sigma_t
% -
% \dot \sigma_t
% \Big)
% \bsx,t) \sigma_t\\
%     &=
%     \frac{\dot{\alpha}_t}{\alpha_t}x + \Big(\frac{\dot{\alpha}_t \sigma_t^2}{\alpha_t} - \dot{\sigma}_t \sigma_t\Big)\bsx,t)
% \end{align}
% From \cref{sec:velocity-score-eq} we have \begin{align}
%     \bv(\bx, t) = \frac{\dot \alpha_t}{\alpha_t} \bx - \lambda_t\sigma_t \bs(\bx, t)
% \end{align}
% This linear equation implies that if $\bs_c$ is the score corresponding to some velocity $\bv_c$, and $\bs_u$ corresponds to velocity $\bv_u$, and both processes share the same $\alpha_t,\sigma_t$, then
% for a velocity $\bv^\zeta$ defined as $\zeta \bv_1 + (1-\zeta)\bv_2$, the score satisfies $\bs^\zeta = \zeta \bs_1 + (1-\zeta)\bs_2$.
% We next show that if $\bs^u = \nabla \log p_t(\bx_t)$ (unconditional score) and $\bs^c = \nabla \log p_t(\bx_t| \by)$ (conditional score),
% then $\bs^\zeta$ is the score for the distribution
% proportional to $p_t(\bx_t)p_t(\by|\bx_t)^\zeta$.
Consider now
\begin{align}
    \bs^\zeta(\bx,t|\by) 
    &\equiv  
        (1-\zeta) \bs(\bx,t) + \zeta \bs(\bx,t|\by)\\
    &=
        \nabla \log p_t(\bx) - \zeta \nabla \log p_t(\bx)
        + \zeta \nabla \log p_t(\bx|\by)\\
    &=
        \nabla \log p_t(\bx) - \zeta \nabla \log p_t(\bx)
        + \Big( \zeta 
            \nabla \log p_t(\by|\bx)
            +
            \zeta \nabla \log p_t(\bx)
        \Big)\\
    &=
        \nabla \log p_t(\bx) 
        + \zeta 
            \nabla \log p_t(\by|\bx)\\
    &=
        \nabla \log [p_t(\bx)  p^\zeta_t(\by|\bx)]
        \label{eq:score:mix}
\end{align}
where we have used the fact $\nabla_{\bx} \log p_t(\bx|\by)= \nabla_{\bx} \log p_t(\by|\bx) + \nabla_{\bx} \log p_t(\bx)$ that follows from $p_t(\bx|\by) p(\by) = p_t(\by|\bx) p_t(\bx)$, and $\zeta$ to be some constant greater than $1$. 
% On the other hand
% \begin{align}
%     \nabla  \log p_t(\bx_t)p_t(\by|\bx_t)^\zeta
%     &=
%     \nabla \log p_t(\bx_t) + \nabla \zeta \log p_t(\by|\bx_t)
%     =
%     \nabla \log p_t(\bx_t) + \zeta \nabla \log p_t(\by|\bx_t)
% \end{align}
\cref{eq:score:mix} shows that using the score mixture $\bs^\zeta(\bx,t|\by) = (1-\zeta) \bs(\bx,t) + \zeta \bs(\bx,t|\by)$, and the velocity mixture associated with it, namely,
\begin{align}
    \bv^\zeta(\bx,t|\by) &= (1-\zeta) \bv(\bx,t) + \zeta \bv(\bx,t|\by)\\
    & = \frac{\dot \alpha_t}{\alpha_t} \bx - \lambda_t\sigma_t [ (1-\zeta) \bs(\bx,t) + \zeta \bs(\bx,t|\by)]\\
    &= \frac{\dot \alpha_t}{\alpha_t} \bx - \lambda_t\sigma_t \bs^\zeta(\bx,t | \by),
     \label{eq:velocity:mixture}
\end{align}
allows one to to construct generative models that sample the tempered distribution $p_t(\bx_t)p^\zeta_t(\by|\bx_t)$ following classifier guidance \cite{dhariwal2021diffusion}.
Note that $p_t(\bx)p^\zeta_t(\by|\bx) \propto p^\zeta_t(\bx | \by) p^{1-\zeta}_t(\bx)$, so we can also perform classifier free guidance sampling \cite{ho2022classifier}. Empirically, we observe significant performance boost by applying classifier free guidance, as showed in \cref{tab:FID-50K} and \cref{tab:FID-50K-IS}. 

\section{Sampling with ODE and SDE}
\label{sec:ode-sde-algo}

\begin{table}
  \centering
  \caption{\textbf{FID-50K scores produced by ODE and SDE.} We demonstrate the comparison between ODE and SDE across all of our model sizes. All statistics are produced without classifier free guidance. Each cell in the table is showing [ODE results] / [SDE results]. We note the better performances of SDE observed in all model sizes are in line with the bounds given in \cite{albergo2023stochastic}, and that ODE has its advantage in lower NFE region, as shown in \cref{fig:nfe_sampler}}
  \scalebox{0.9}{
  \begin{tabular}{lcccccc}
    \toprule
    Model  & Training Steps(K) &FID$\downarrow$ & sFID$\downarrow$ & IS$\uparrow$ & Precision$\uparrow$ & Recall$\uparrow$ \\
    \midrule
    SiT-S & 400 & 58.97 / 57.64 & 8.95 / 9.05 & 23.34 / 24.78 & 0.40 / 0.41 & 0.59 / 0.60\\
    \midrule 
    SiT-B & 400 & 34.84 / 33.02 & 6.59 / 6.46 & 41.53 / 43.71 & 0.52 / 0.53 & 0.64 / 0.63\\
    \midrule
    SiT-L & 400 & 20.01 / 18.79 & 5.31 / 5.29 & 67.76 / 72.02 & 0.62 / 0.64 & 0.64 / 0.64\\
    \midrule
    SiT-XL & 400 & 18.04 / 17.19 & 5.17 / 5.07 & 73.90 / 76.52 & 0.63 / 0.65 & 0.64 / 0.63\\
    SiT-XL & 7000  & 9.35 / 8.26 & 6.38 / 6.32 & 126.06 / 131.65 & 0.67 / 0.68 & 0.68 / 0.67\\
    % SiT-XL $_{\text{(cfg=1.5)}}$ & 7000 & 2.15 / 2.06 & 4.60 / 4.50 & 258.09 / 270.27 & 0.81 / 0.82 & 0.60 / 0.59 \\ 
    % \textbf{SiT-XL(cfg = 1.5, ODE)}   & 2.15  & 4.60 & 258.09 & 0.81 & 0.60 \\
    % \textbf{SiT-XL(cfg = 1.5, SDE:$\sigma_t$)}   & \textbf{2.06} & 4.50 &  270.27 & 0.82 & 0.59\\
    \arrayrulecolor{black}\bottomrule
  \end{tabular}
  }
  \label{tab:FID-50K-ODE-SDE}
\end{table}

In the main body of the paper, we used a second order Heun integrator for solving the ODE in~\cref{eq:prob:flow:ode} and a first order Euler-Maruyama integrator for solving the SDE in~\cref{eq:sde}. We summarize all results in \cref{tab:FID-50K-ODE-SDE}, and present the implementations below.

It is feasible to use either a velocity model $\bv_\theta$ or a score model $\bs_\theta$ in applying the above two samplers. If learning the score for the deterministic Heun sampler, we could always convert the learned $\bs_\theta$ to $\bv_\theta$ following \cref{sec:velocity-score-eq}. However, as there exists potential numerical instability (depending on interpolants) in $\dot \sigma_t$, $\alpha_t^{-1}$ and $\lambda_t$, it's recommended to learn $\bv_\theta$ in sampling with deterministic sampler instead of $\bs_\theta$. For the stochastic sampler, it's required to have both $\bv_\theta$ and $\bs_\theta$ in integration, so we always need to convert from one (either learning velocity or score) to obtain the other. Under this scenario, the numerical issue from \cref{sec:velocity-score-eq} can only be avoided by clipping the time interval near $t = 0$. Empirically we found clipping the interval by $h = 0.04$ and doing a long last step from $t = 0.04$ to $0$ can greatly benefit the performance. A detailed summary of sampler configuration is provided in \cref{sec:imple}. 

Additionally, we could replace $\bv_\theta$ and $\bs_\theta$ by $\bv_\theta^\zeta$ and $\bs_\theta^\zeta$ presented in \cref{sec:guidance} as inputs of the two samplers and enjoy the performance improvements coming along with guidance. As guidance requires evaluating both conditional and unconditional model output in a single step, it will impose twice the computational cost when sampling.

We also note that our models are compatible with more advanced samplers~\cite{kingma2023variational, lu2022dpmsolver}. We do not include the evaluations of those samplers in our work for the sake of direct comparison with the DDPM model, and we leave the investigation of potential performance improvements to future work. 

\paragraph{Comparison between DDPM and Euler-Maruyama}
We primarily investigate and report the performance comparison between DDPM and Euler-Maruyama samplers. We set our Euler sampler's number of steps to be 250 to match that of DDPM during evaluation. This comparison is made direct and fair, as the DDPM method can be viewed as a discretized version of Euler's method. 

\begin{figure}[h!]
  \centering
  \vspace{-2em}
   \includegraphics[width=0.24\linewidth]{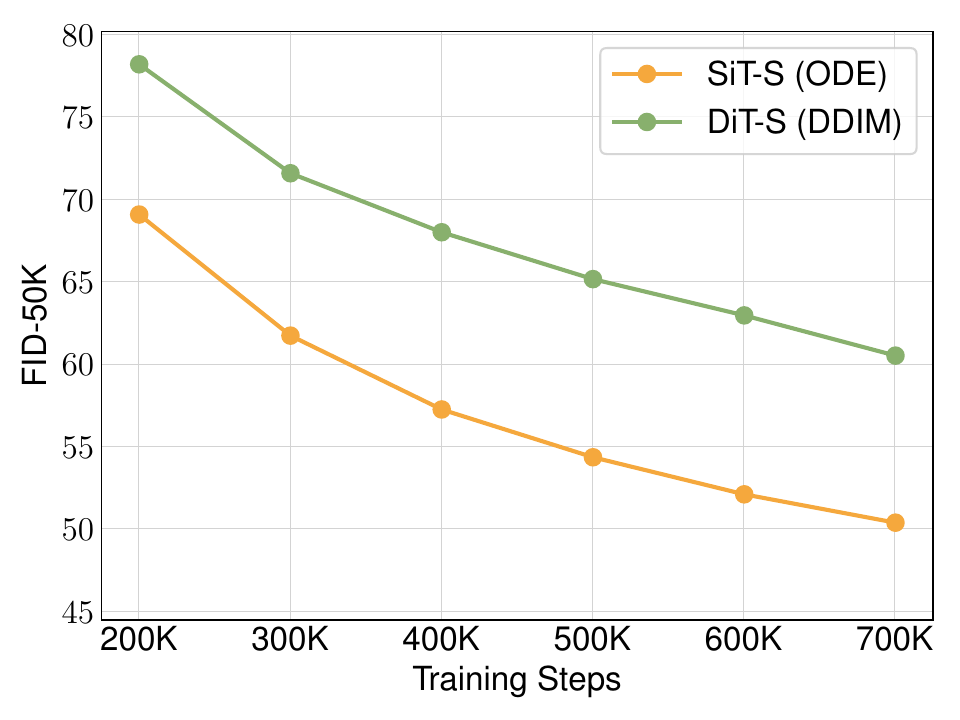}
   \includegraphics[width=0.24\linewidth]{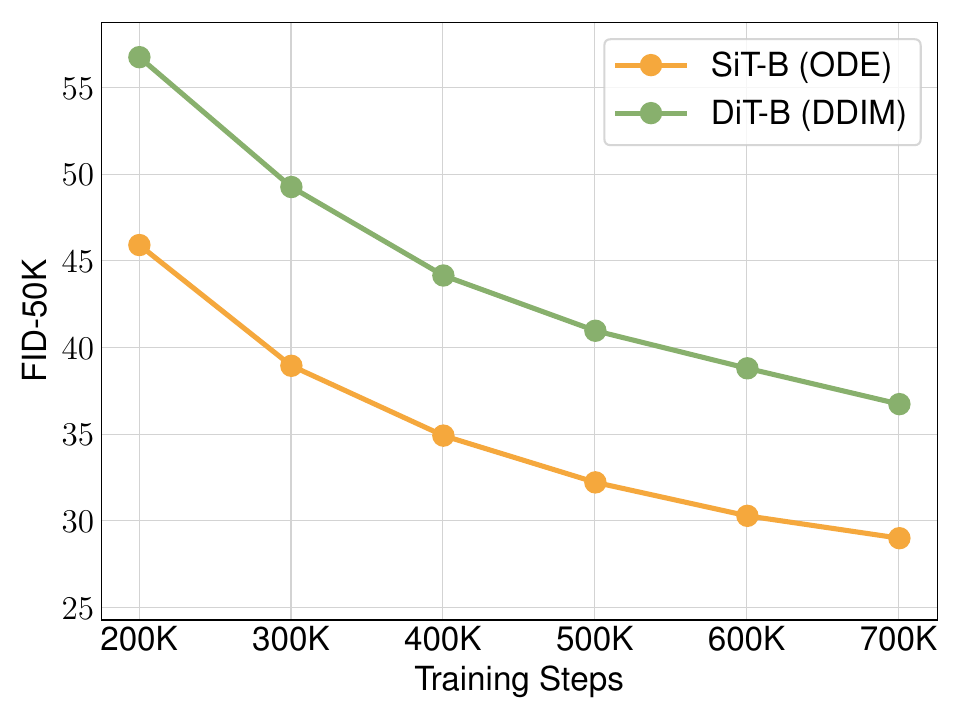}
   \includegraphics[width=0.24\linewidth]{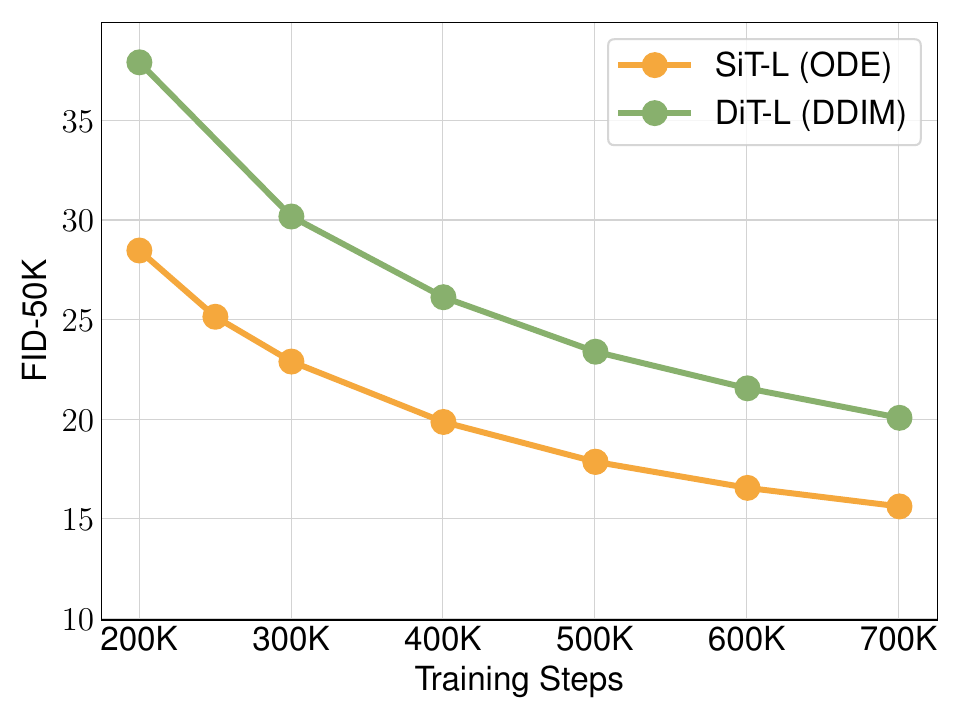}
   \includegraphics[width=0.24\linewidth]{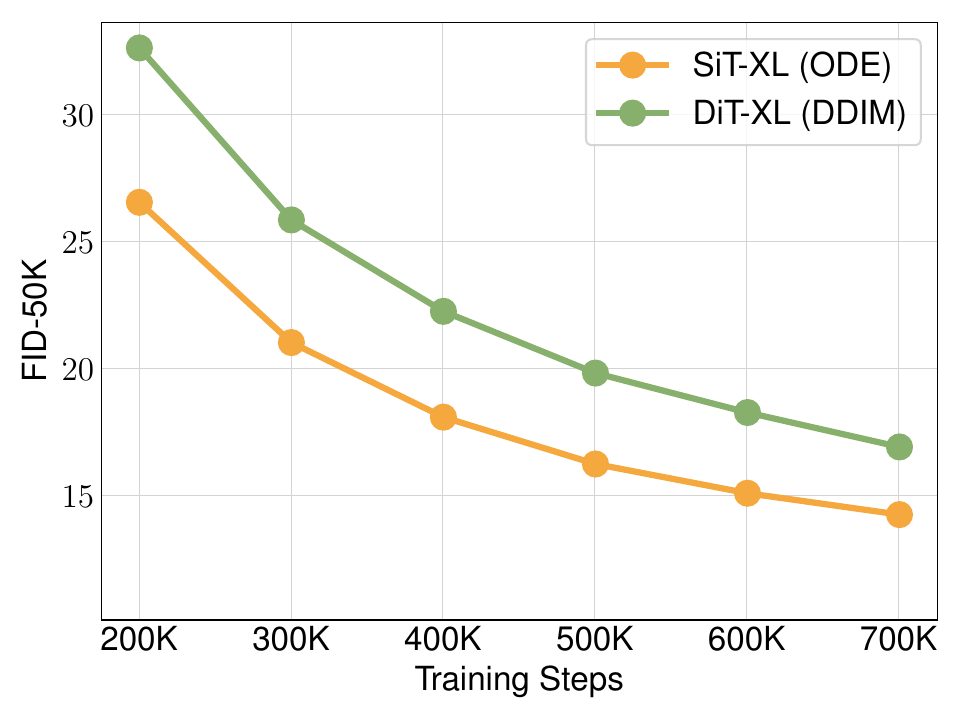}
    \caption{\textbf{SiT observes improvement in FID across all model sizes.} We show FID-50K over training iterations for both DiT and SiT models. Across all model sizes, SiT converges faster. We acknowledge this is \textit{not directly an apples-to-apples comparison}. This is because DDIM is essentially a discrete form of the first-order Euler's method, whereas in sampling SiT, we employ the second-order Heun’s method. Nevertheless, both the SiT and DiT results are produced by a deterministic sampler with a 250 NFE.
       }
   \label{fig:FID-50K-ode}
\end{figure}

\paragraph{Comparison between DDIM and Heun}
We also investigate the performance difference produced by deterministic samplers between DiT and our models. In \cref{fig:FID-50K-ode}, we show the FID-50K results for both DiT models sampled with DDIM and SiT models sampled with Heun. We note that this is not directly an apples-to-apples comparison, as DDIM can be viewed as a discretized version of the first order Euler's method, while we use the second order Heun's method in sampling SiT models, due to the large discretization error with Euler's method in continuous time. Nevertheless, we control the NFEs for both DDIM (250 sampling steps) and Heun (250 NFE). %We include the results here for completeness of our discussion. 

\begin{figure}[h]
    \centering
    \vspace{-4em}
    \begin{minipage}{.7\linewidth}
        \begin{algorithm}[H]
        \label{alg:ODE-sampler}
            \caption{Deterministic Heun Sampler}
            \begin{algorithmic}
                \Procedure{HeunSampler}{$\bv_\theta(\bx, t, \by), t_{i \in \{0,\cdots,N\}}, \alpha_t, \sigma_t$}
                    \State \textbf{sample} $\bx_0 \sim \NN(0, \mathbf{I})$ \Comment{Generate initial sample}
                    \State $\Delta t \gets t_1 - t_0$\Comment{Determine fixed step size}
                    \For{$i \in \{0, \cdots, N-1\}$}
                        \State $\mathbf{d}_i \gets \bv_\theta(\bx_i, t_i, \by)$
                        \State $\tilde{\bx}_{i+1} \gets \bx_i + \Delta t \mathbf{d}_i$ \Comment{Euler Step at $t_i$}
                        \State $\mathbf{d}_{i+1} \gets \bv_\theta(\tilde{\bx}_{i+1}, t_{i+1}, \by)$
                        \State $\bx_{i+1} \gets \bx_i + \frac{\Delta t}{2} [\mathbf{d}_i + \mathbf{d}_{i+1}]$ \Comment{Explicit trapezoidal rule at $t_{i+1}$}
                    \EndFor
                    \State \Return $\bx_N$
                \EndProcedure
            \end{algorithmic}
        \end{algorithm}
    \end{minipage}
    \vspace{-2em}
\end{figure}

\begin{figure}
\label{algo:EM}
\centering
\vspace{-3em}
    \begin{minipage}{.6\linewidth}
        \begin{algorithm}[H] 
            \caption{Stochastic Euler-Maruyama Sampler}
            \begin{algorithmic}
                \Procedure{EulerSampler}{$\bv_\theta(\bx, t, \by), w_t, t_{i \in \{0,\cdots,N\}}, T, \alpha_t, \sigma_t$}
                    \State \textbf{sample} $\bx_0 \sim \NN(0, \mathbf{I})$ \Comment{Generate initial sample}
                    \State $\bs_\theta \gets$ \textbf{convert} from $\bv_\theta$ following \cref{sec:velocity-score-eq}\Comment{Obtain $\nabla_\bx \log p_t(\bx)$ in \cref{eq:sde}}
                    \State $\Delta t \gets t_1 - t_0$ \Comment{Determine fixed step size}
                    \For{$i \in \{0, \cdots, N-1\}$}
                        \State \textbf{sample} $\eps_i \sim \NN(0, \mathbf{I})$ 
                        \State $\rd \eps_i \gets \eps_i * \sqrt{\Delta t}$
                        \State $\mathbf{d}_i \gets \bv_\theta(\bx_i, t_i, \by) + \frac{1}{2} w_{t_i} \bs_\theta(\bx_i, t_i, \by)$ \Comment{Evaluate drift term at $t_i$}
                        \State $\bar{\bx}_{i+1} \gets \bx_i + \Delta t \mathbf{d}_i$
                        \State $\bx_{i+1} \gets \bar{\bx}_{i+1} + \sqrt{w_{t_i}}\rd \eps_i $ \Comment{Evaluate diffusion term at $t_i$}
                    \EndFor
                    \State $h \gets T - t_N$ \Comment{Last step size; $T$ denotes the time where $\bx_T = \bx_*$}
                    \State $\mathbf{d} \gets \bv_\theta(\bx_N, t_N, \by) + \frac{1}{2} w_{t_N} \bs_\theta(\bx_N, t_N, \by)$
                    \State $\bx \gets \bx_{N} + h * \mathbf{d}$ \Comment{Last step; output noiseless sample without diffusion}
                    \State \Return $\bx$
                \EndProcedure
            \end{algorithmic}
        \end{algorithm}
    \end{minipage}
    \vspace{-2em}
\end{figure}

\section{Additional Implementation Details}
\label{sec:imple}
We implemented our models in JAX following the DiT PyTorch codebase by \cite{peebles2023scalable}\footnote{https://github.com/facebookresearch/DiT}, and referred to \cite{albergo2023stochastic}\footnote{https://github.com/malbergo/stochastic-interpolants}, \cite{song2021scorebased}\footnote{https://github.com/yang-song/score\_sde}, and \cite{dockhorn2022scorebased}\footnote{https://github.com/nv-tlabs/CLD-SGM} for our implementation of the Euler-Maruyama sampler. For the Heun sampler, we directly used the one from \texttt{diffrax}~\cite{kidger2021on}\footnote{https://github.com/patrick-kidger/diffrax}, a JAX-based numerical differential equation solver library. 

\paragraph{Architectural Configurations} 
We follow the identical transformer architectures in DiT and have four different configurations: SiT-\{S,B,L,XL\}, varying in model size (parameters) and compute (flops). A detailed summarization is presented below. 

\begin{table}
\centering
\small
\caption{\textbf{Details of SiT models.}
We follow DiT~\cite{peebles2023scalable} for the Small (S), Base (B), Large (L) and XLarge (XL) model configurations.}
\begin{tabular}{l c c c c c}
\toprule
Model            & Layers $N$ & Hidden size $d$ &  Heads   \\
\midrule 
SiT-S   &   12   &     384   &   6    \\
SiT-B   &   12   &      768    &   12    \\
SiT-L  &    24   &      1024    &   16    \\
SiT-XL &    28  &       1152     &   16   \\
\bottomrule
\end{tabular}
\label{tbl:models}
\end{table}

\paragraph{Training configurations} 
We trained all of our models following identical structure and hyperparameters retained from DiT \cite{peebles2023scalable}. We used AdamW~\cite{kingma2017adam, loshchilov2019decoupled} as optimizer for all models. We use a constant learning rate of $1\times10^{-4}$ and a batch size of $256$. We used random horizontal flip with probability of $0.5$ in data augmentation. \textit{We did not tune the learning rates, decay/warm up schedules, AdamW parameters, nor use any extra data augmentation or gradient clipping during training}. Our largest model, SiT-XL, trains at approximately $6.8$ iters/sec on a TPU v4-64 pod following the above configurations. This speed is slightly faster compared to DiT-XL, which trains at  ~$6.4$ iters/sec under identical settings. 

\paragraph{Sampling configurations} We maintain an exponential moving average (EMA) of all models weights over training with a decay of $0.9999$. All results are sampled from the EMA checkpoints, which is empirically observed to yield better performance. We summarize the start and end points of our deterministic and stochastic samplers with different interpolants below, where each $t_0$ and $t_N$ are carefully tuned to optimize performance and avoid numerical instability during integration.

\begin{table}[!h]
    \centering
    \caption{Sampler configurations}
    \begin{tabular}{ccccccc}
    \toprule
     \textit{Interpolant} & \textit{Model} & \textit{Objective} & \multicolumn{2}{c}{Heun} & \multicolumn{2}{c}{Euler-Maruyama} \\
     
      & & &$t_0$ & $t_N$ &$t_0$ & $t_N$ \\
    \midrule
     SBDM-VP & velocity & $\LL_{\mathrm{v}}$ &\texttt{1} & \texttt{1e-5} &\texttt{1} & \texttt{4e-2} \\
     & score & $\LL_{\mathrm{s}_\lambda}$ &\texttt{1} & \texttt{1e-5} &\texttt{1} & \texttt{4e-2} \\
     \midrule
     GVP & velocity & $\LL_{\mathrm{v}}$ & \texttt{1} & \texttt{0} & \texttt{1} & \texttt{4e-2}\\
     & score & $\LL_{\mathrm{s}}$ & \texttt{1 - 1e-5} & \texttt{0} & \texttt{1 - 1e-3} & \texttt{4e-2}\\
     \midrule
     LIN & velocity & $\LL_{\mathrm{v}}$ & \texttt{1} & \texttt{0} & \texttt{1} & \texttt{4e-2}\\
     & score & $\LL_{\mathrm{s}}$ & \texttt{1 - 1e-5} & \texttt{0} & \texttt{1 - 1e-3} & \texttt{4e-2}\\
     \bottomrule
    \end{tabular}
    \label{tab:sampler-config}
\end{table}

\paragraph{FID calculation} We calculate FID scores between generated images (10K or 50K) and all available real images in ImageNet training dataset. We observe small performance variations between TPU-based FID evaluation and GPU-based FID evaluation (ADM's TensorFlow evaluation suite~\cite{dhariwal2021diffusion}\footnote{https://github.com/openai/guided-diffusion/tree/main/evaluations}). To ensure consistency with the basline DiT, we sample all of our models on GPU and obtain FID scores using the ADM evaluation suite. 

\clearpage

\section{Additional Visual results}
\label{sec:add_samples}

\begin{figure}[h]
  \centering
  \begin{minipage}{.45\linewidth}
      \centering
      \includegraphics[width=\linewidth]{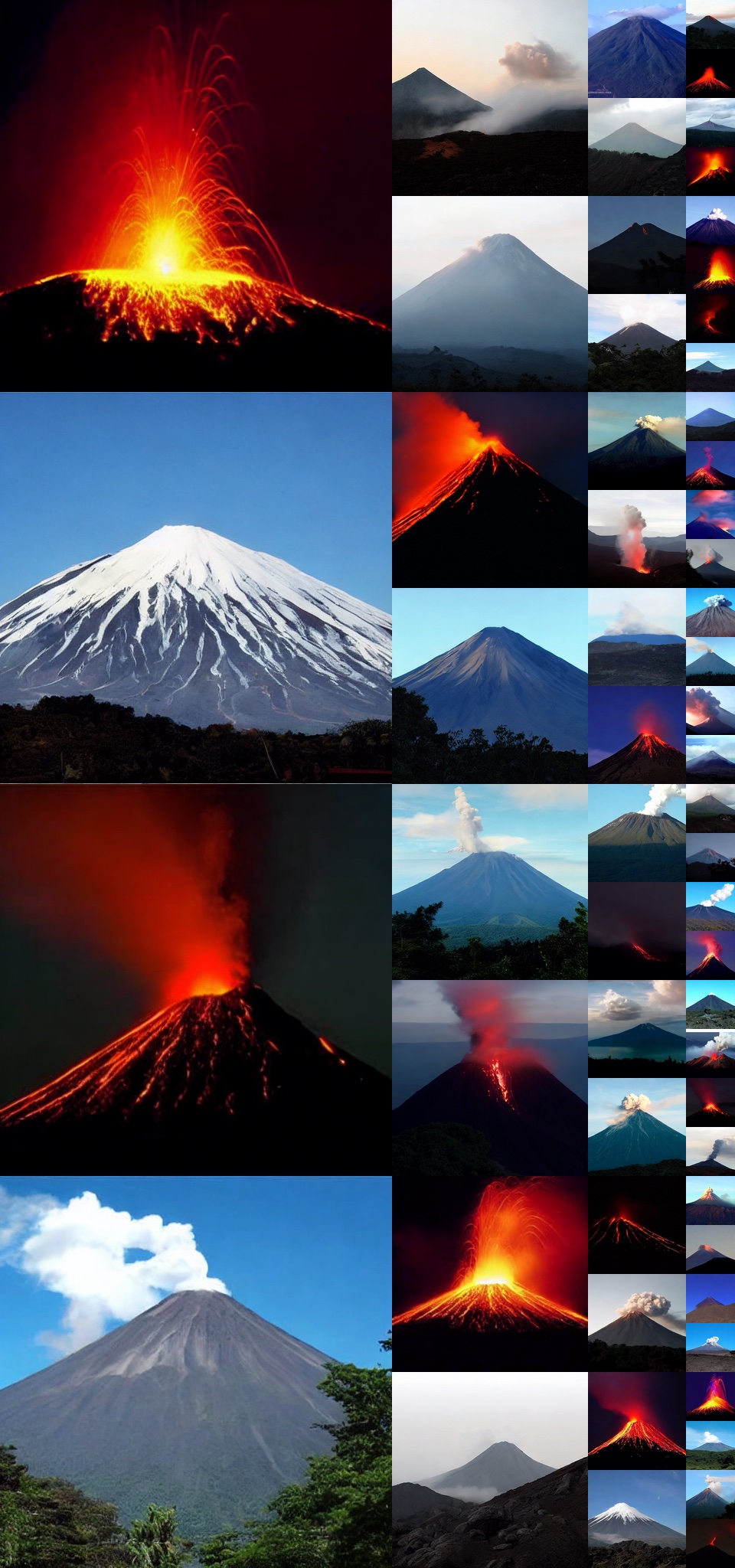}
      \caption[]{\textbf{Uncurated $512\times512$ SiT-XL samples}.\\
           Classifier-free guidance scale = 4.0 \\
           Class label = "volcano"(980)
        }
        \label{fig:superimage-980}
  \end{minipage}
  \hfill
   \begin{minipage}{.45\linewidth}
      \centering
      \includegraphics[width=\linewidth]{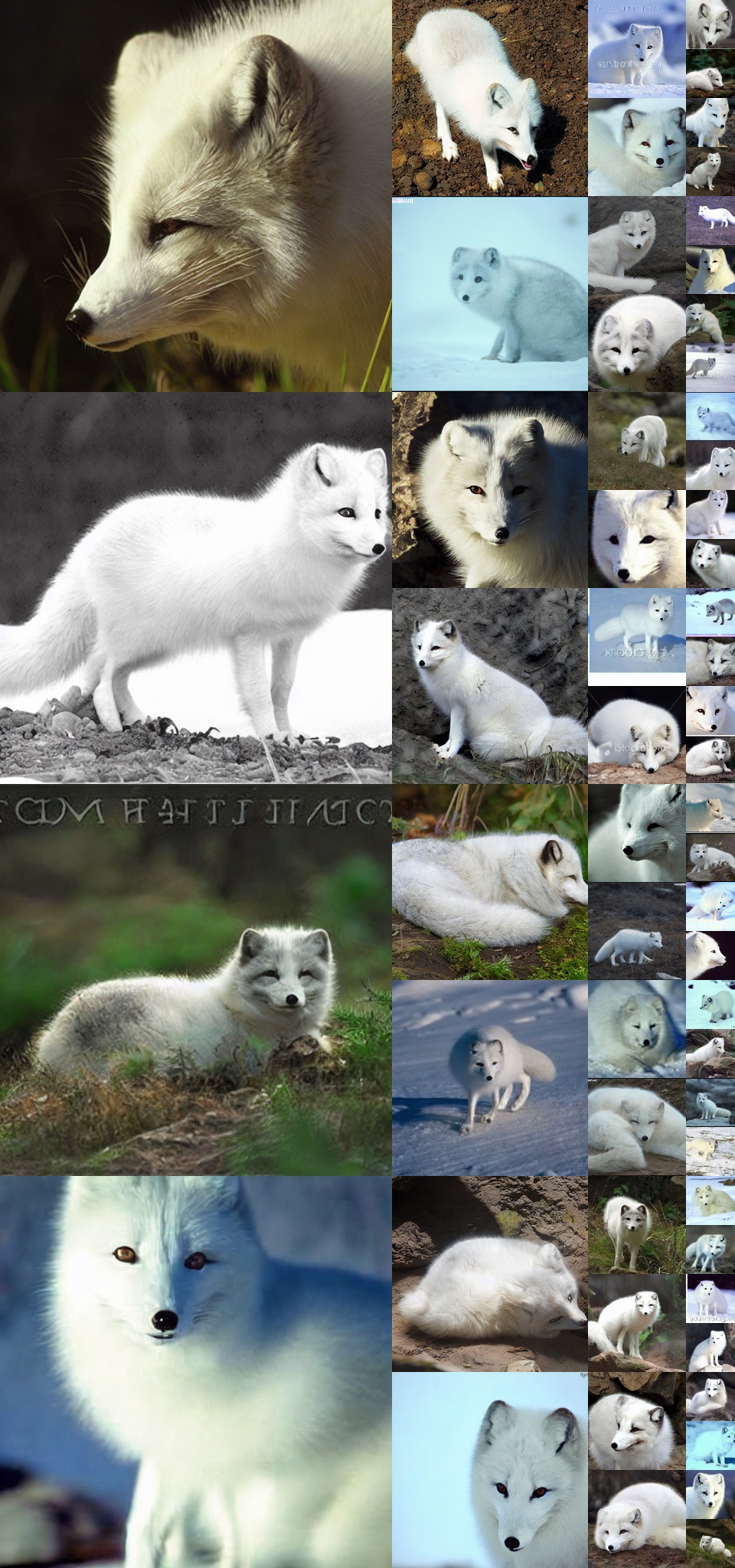}
      \caption[]{\textbf{Uncurated $512\times512$ SiT-XL samples}.\\
           Classifier-free guidance scale = 4.0 \\
           Class label = "arctic fox"(279)
        }
        \label{fig:superimage-279}
  \end{minipage}   
   
\end{figure}

\begin{figure}[h]
  \centering
   \begin{minipage}{.45\linewidth}
      \centering
      \includegraphics[width=\linewidth]{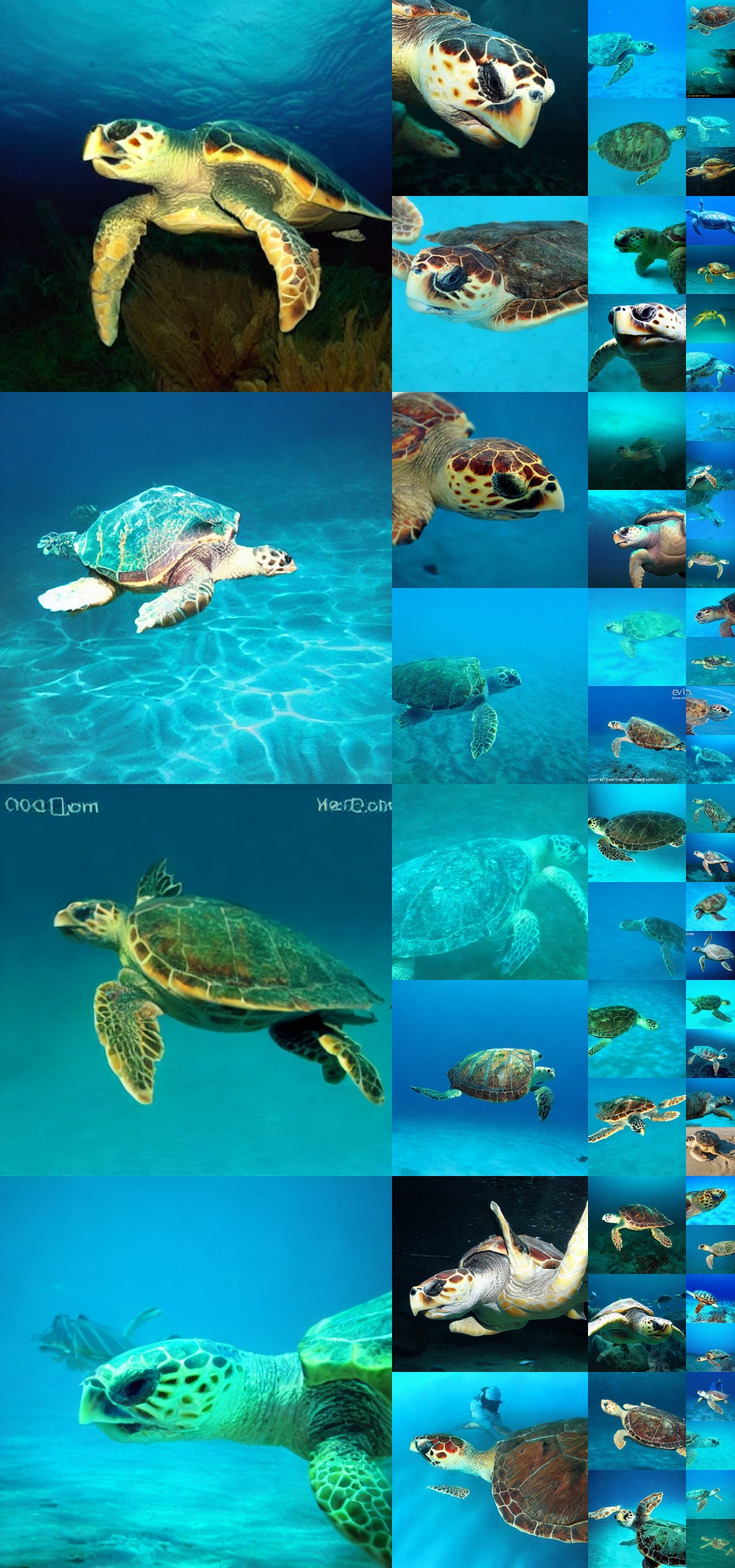}
      \caption[]{\textbf{Uncurated $512\times512$ SiT-XL samples}.\\
           Classifier-free guidance scale = 4.0 \\
           Class label = "loggerhead turtle"(33)
        }
        \label{fig:superimage-33}
  \end{minipage}   
  \hfill
  \begin{minipage}{.45\linewidth}
      \centering
      \includegraphics[width=\linewidth]{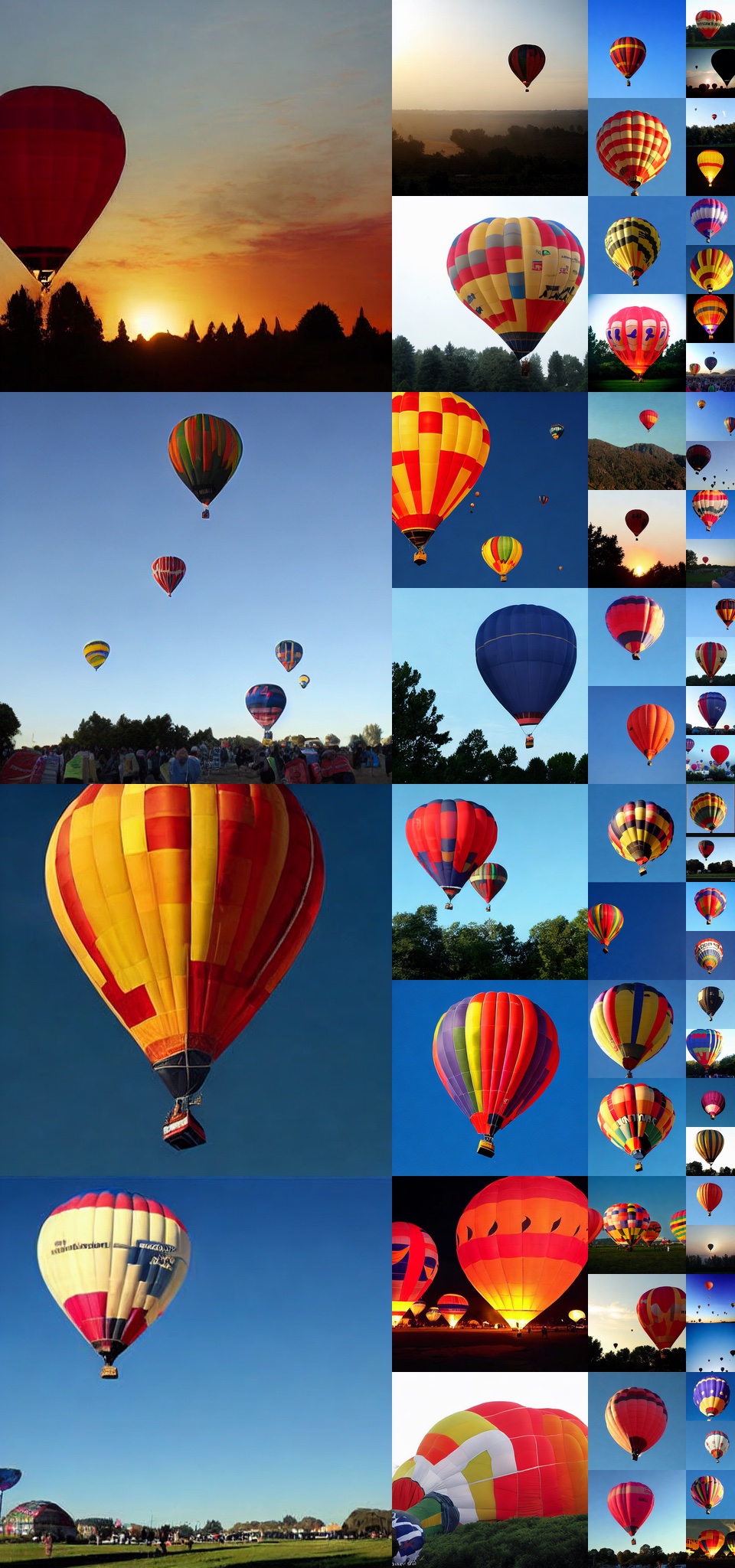}
      \caption[]{\textbf{Uncurated $512\times512$ SiT-XL samples}.\\
           Classifier-free guidance scale = 4.0 \\
           Class label = "balloon"(417)
        }
        \label{fig:superimage-417}
  \end{minipage}
\end{figure}

\begin{figure}[h]
  \centering
   \begin{minipage}{.45\linewidth}
      \centering
      \includegraphics[width=\linewidth]{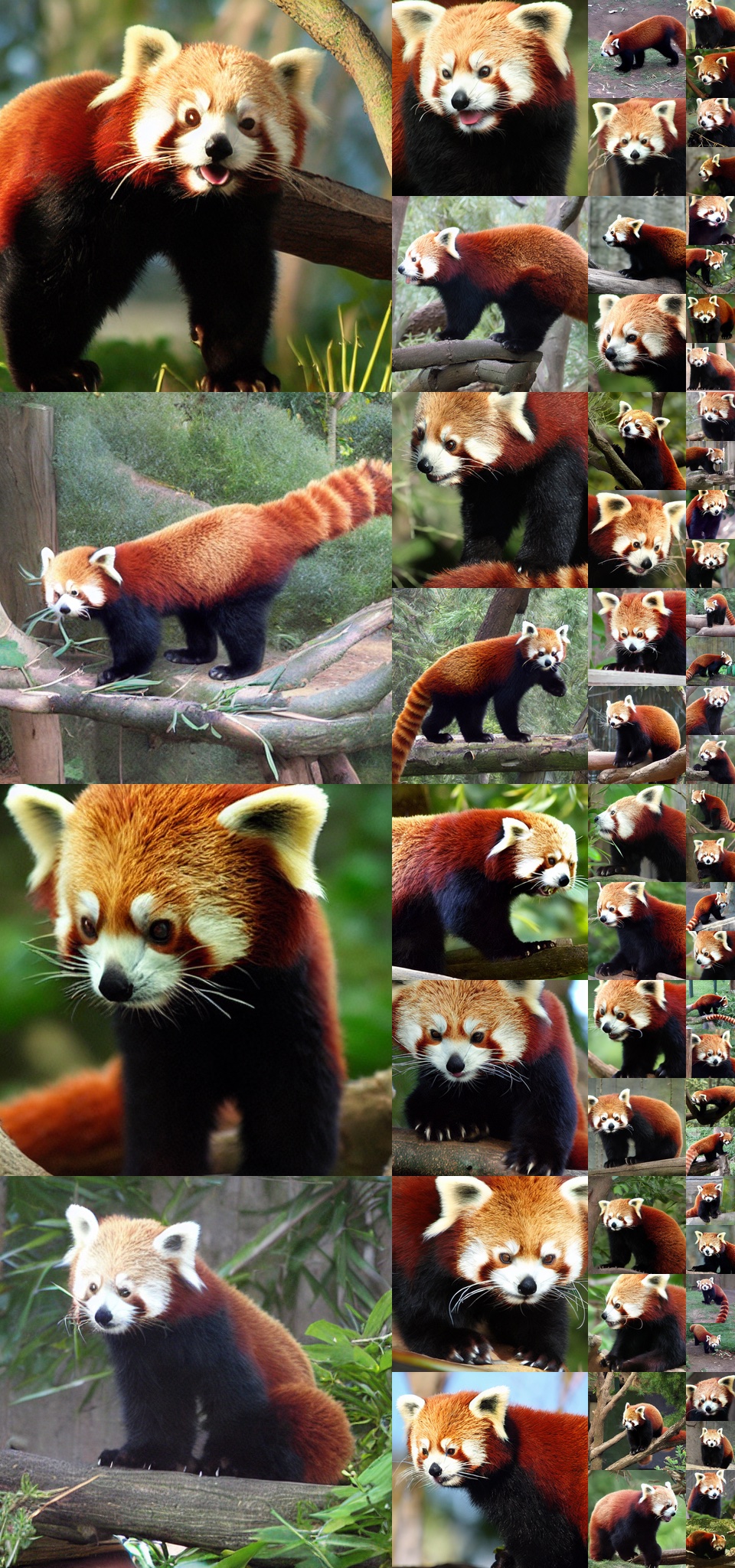}
      \caption[]{\textbf{Uncurated $512\times512$ SiT-XL samples}.\\
           Classifier-free guidance scale = 4.0 \\
           Class label = "red panda"(387)
        }
        \label{fig:superimage-387}
  \end{minipage}   
  \hfill
  \begin{minipage}{.45\linewidth}
      \centering
      \includegraphics[width=\linewidth]{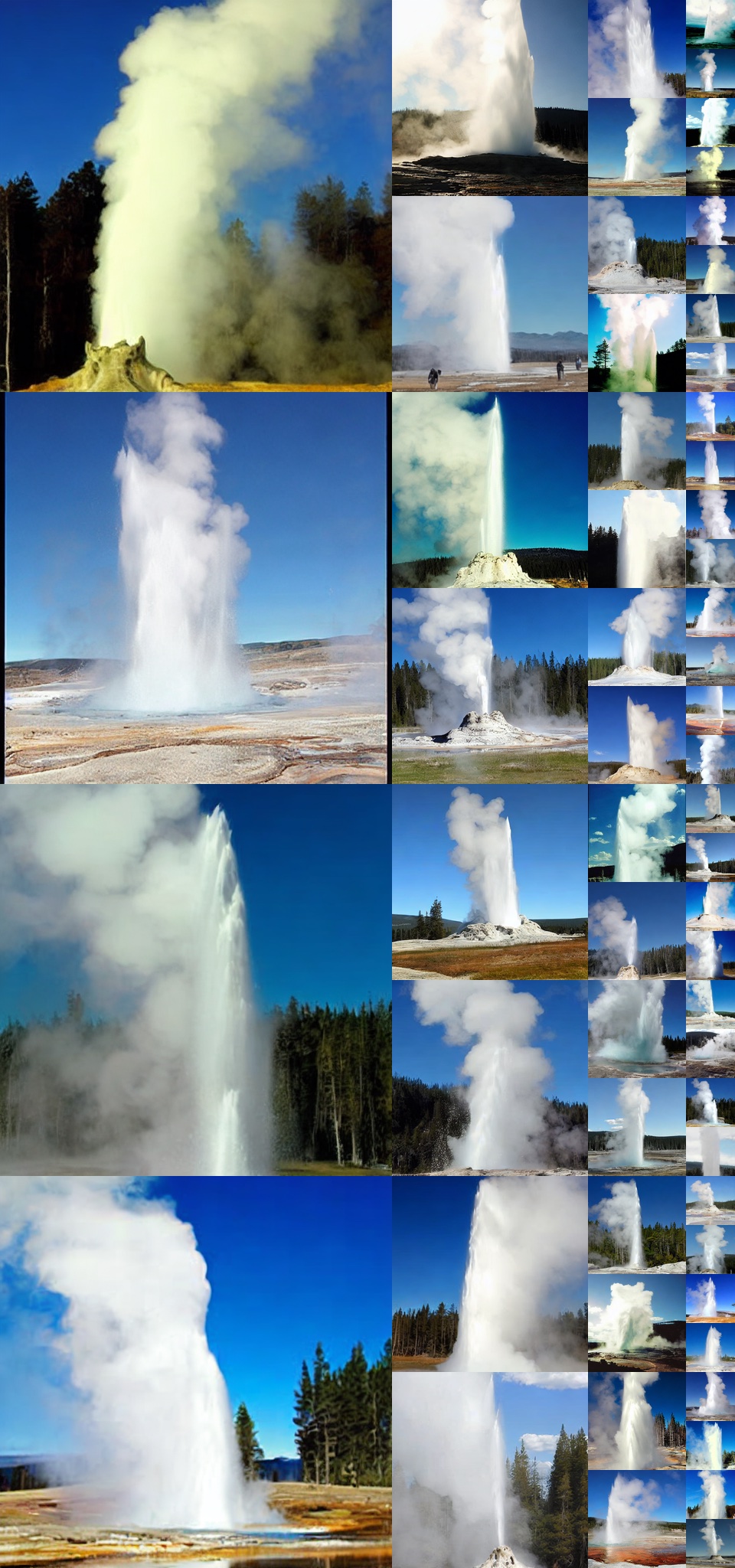}
      \caption[]{\textbf{Uncurated $512\times512$ SiT-XL samples}.\\
           Classifier-free guidance scale = 4.0 \\
           Class label = "geyser"(974)
        }
        \label{fig:superimage-974}
  \end{minipage}
\end{figure}

\begin{figure}[h]
  \centering
  \begin{minipage}{.45\linewidth}
      \centering
      \includegraphics[width=\linewidth]{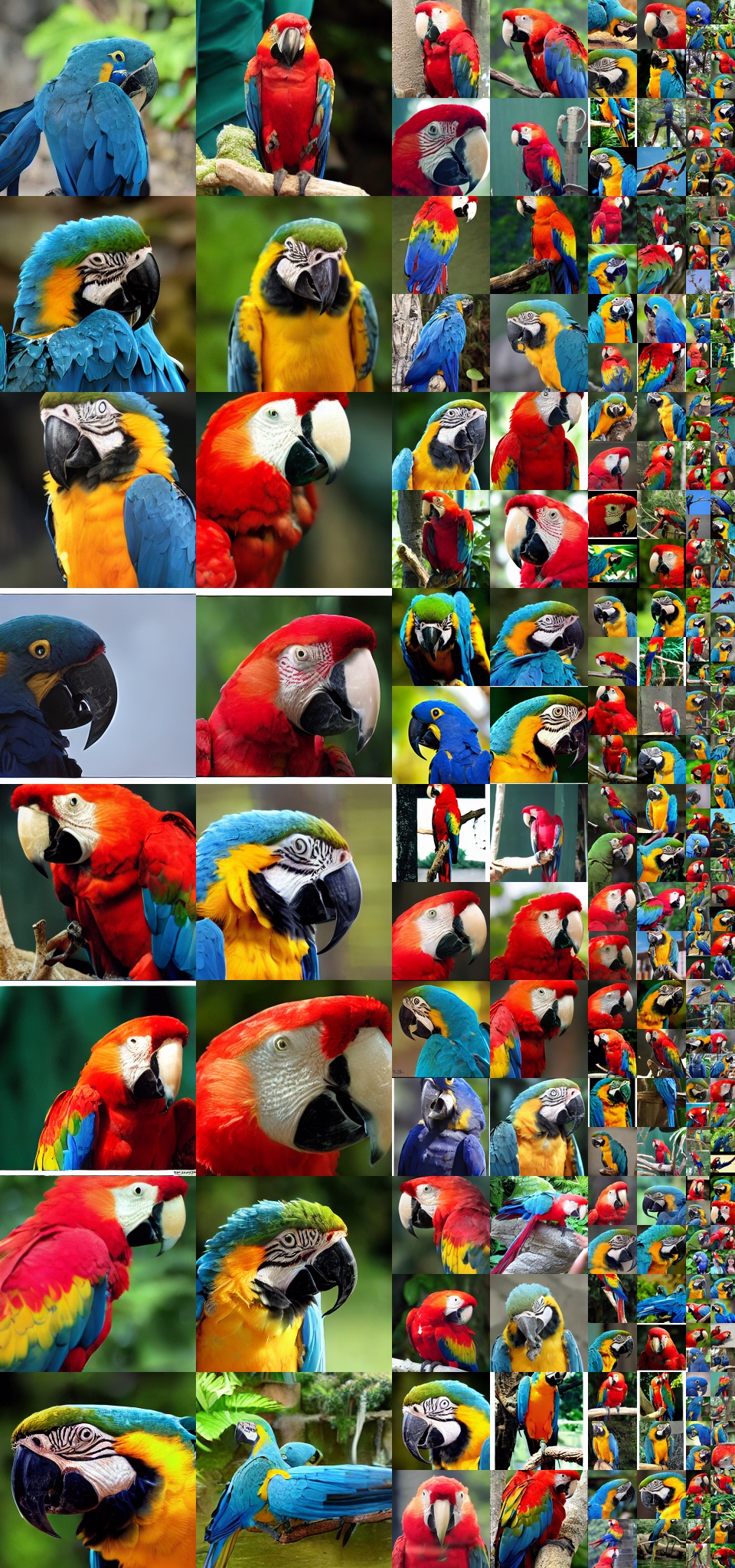}
      \caption[]{\textbf{Uncurated $256\times256$ SiT-XL samples}.\\
           Classifier-free guidance scale = 4.0 \\
           Class label = "macaw"(88)
        }
        \label{fig:superimage-88}
  \end{minipage}
  \hfill
  \hfill
   \begin{minipage}{.45\linewidth}
      \centering
      \includegraphics[width=\linewidth]{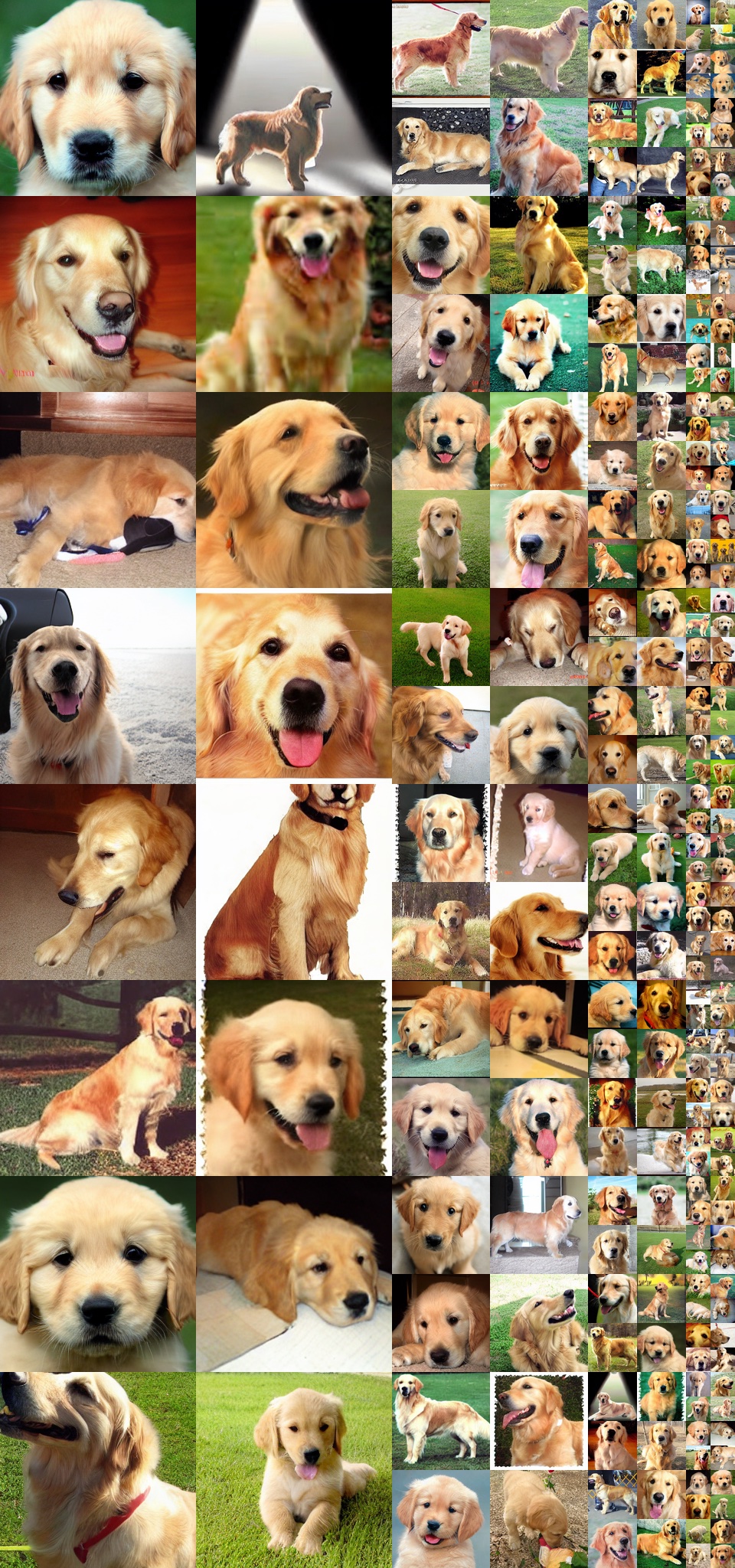}
      \caption[]{\textbf{Uncurated $256\times256$ SiT-XL samples}.\\
           Classifier-free guidance scale = 4.0 \\
           Class label = "golden retriever"(207)
        }
        \label{fig:superimage-207}
  \end{minipage}   
   
\end{figure}

\begin{figure}[h]
  \centering
  \begin{minipage}{.45\linewidth}
      \centering
      \includegraphics[width=\linewidth]{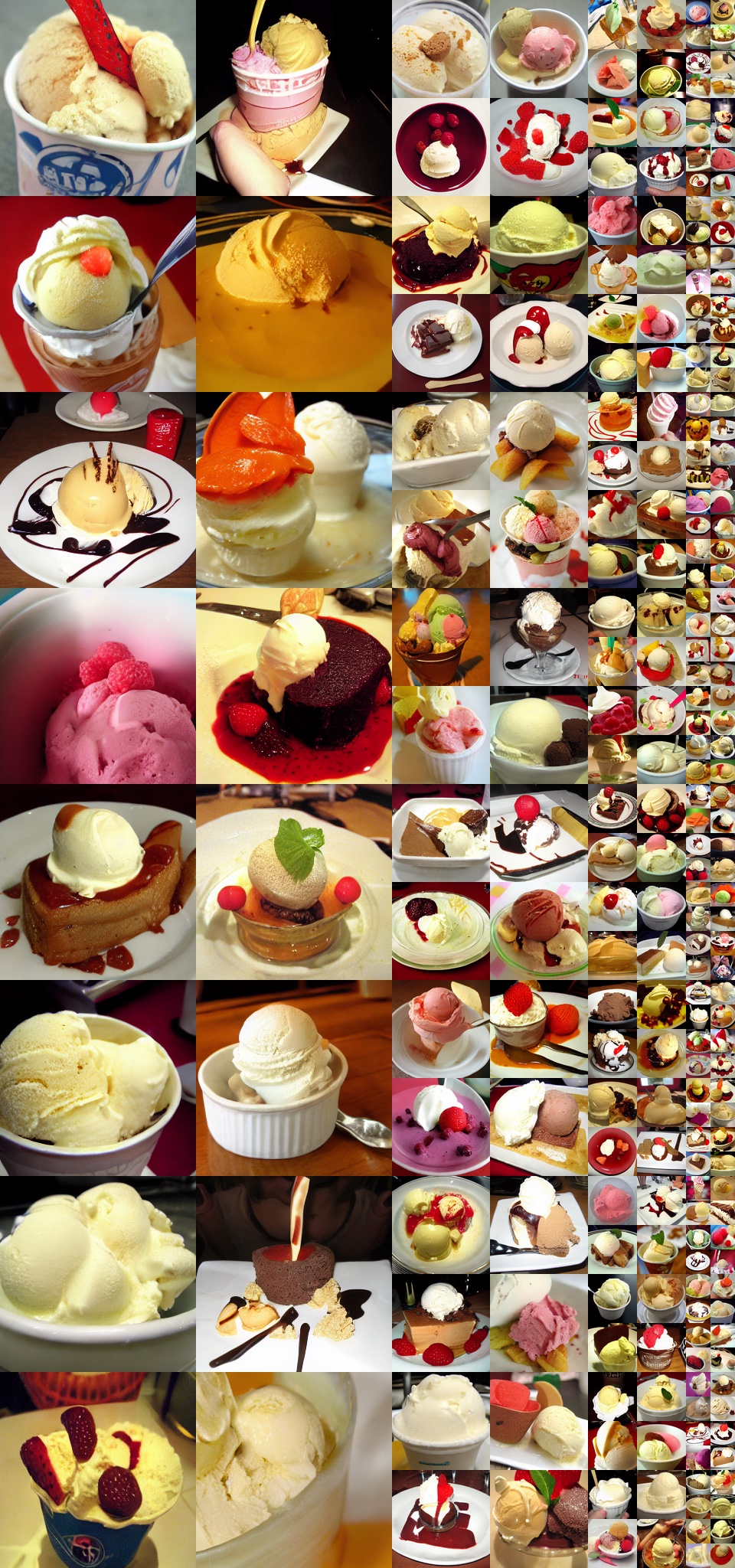}
      \caption[]{\textbf{Uncurated $256\times256$ SiT-XL samples}.\\
           Classifier-free guidance scale = 4.0 \\
           Class label = "	ice cream"(928)
        }
        \label{fig:superimage-928}
  \end{minipage}
  \hfill
   \begin{minipage}{.45\linewidth}
      \centering
      \includegraphics[width=\linewidth]{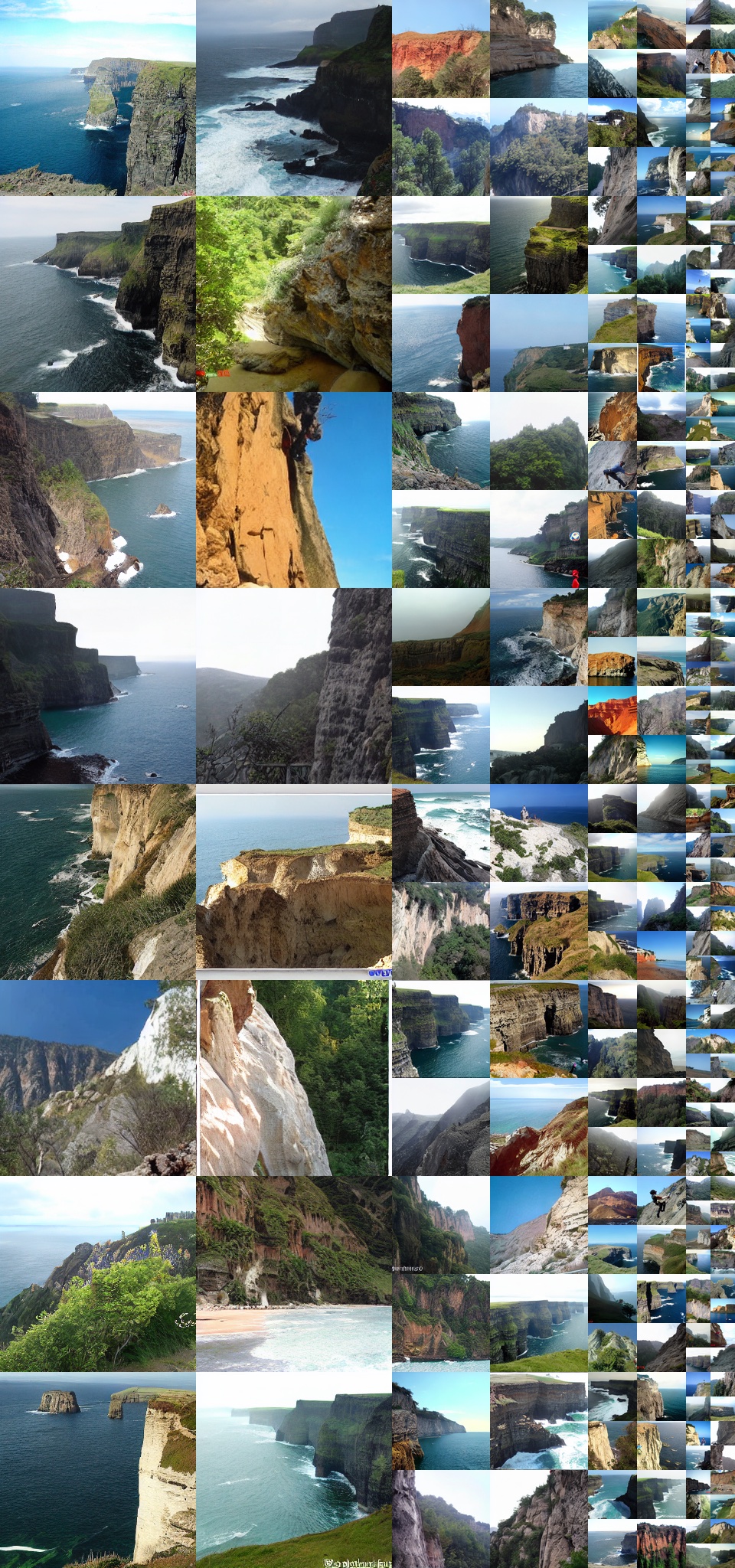}
      \caption[]{\textbf{Uncurated $256\times256$ SiT-XL samples}.\\
           Classifier-free guidance scale = 4.0 \\
           Class label = "cliff"(972)
        }
        \label{fig:superimage-972}
  \end{minipage}   
   
\end{figure}

\begin{figure}[h]
  \centering
   \begin{minipage}{.45\linewidth}
      \centering
      \includegraphics[width=\linewidth]{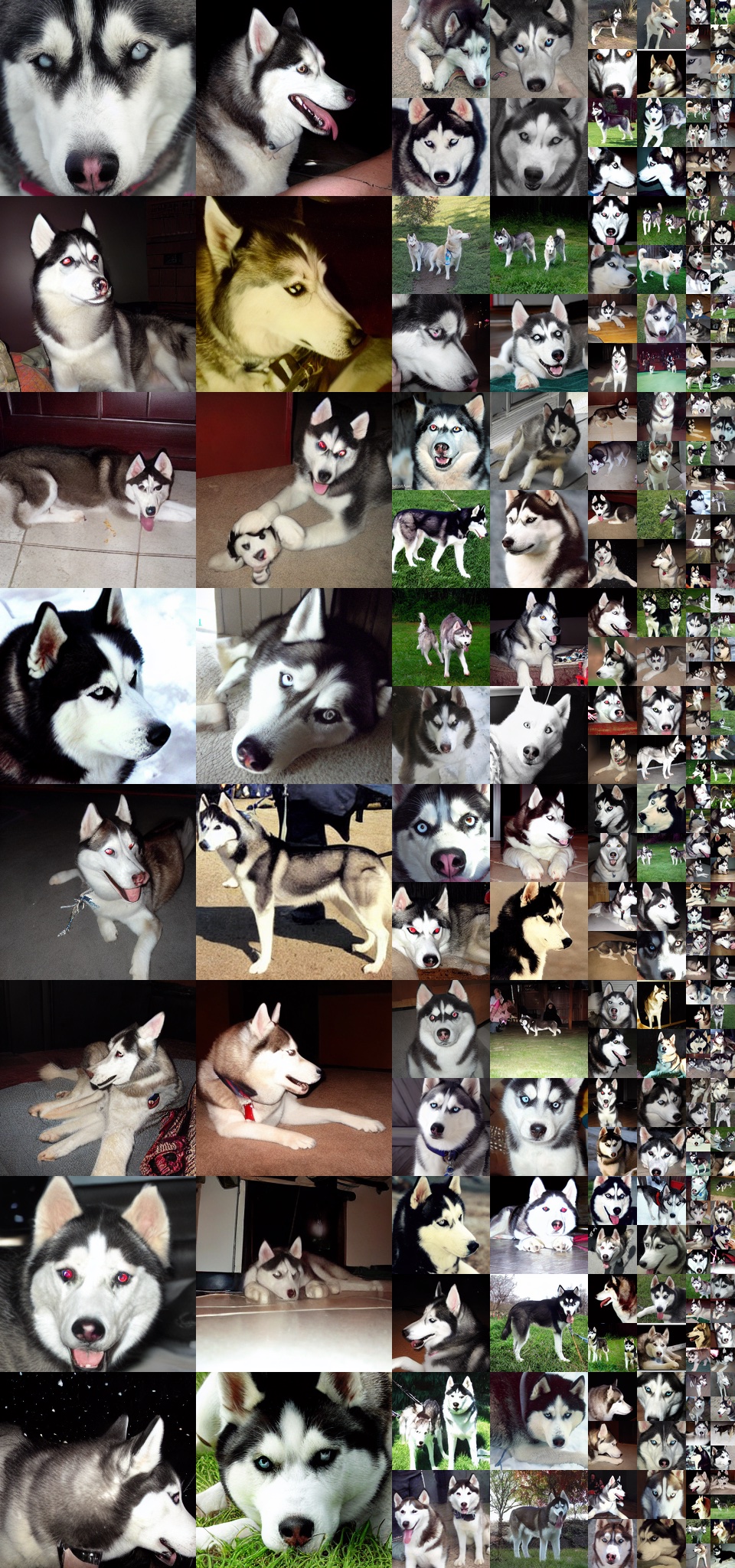}
      \caption[]{\textbf{Uncurated $256\times256$ SiT-XL samples}.\\
           Classifier-free guidance scale = 4.0 \\
           Class label = "husky"(250)
        }
        \label{fig:superimage-250}
  \end{minipage}   
  \hfill
  \begin{minipage}{.45\linewidth}
      \centering
      \includegraphics[width=\linewidth]{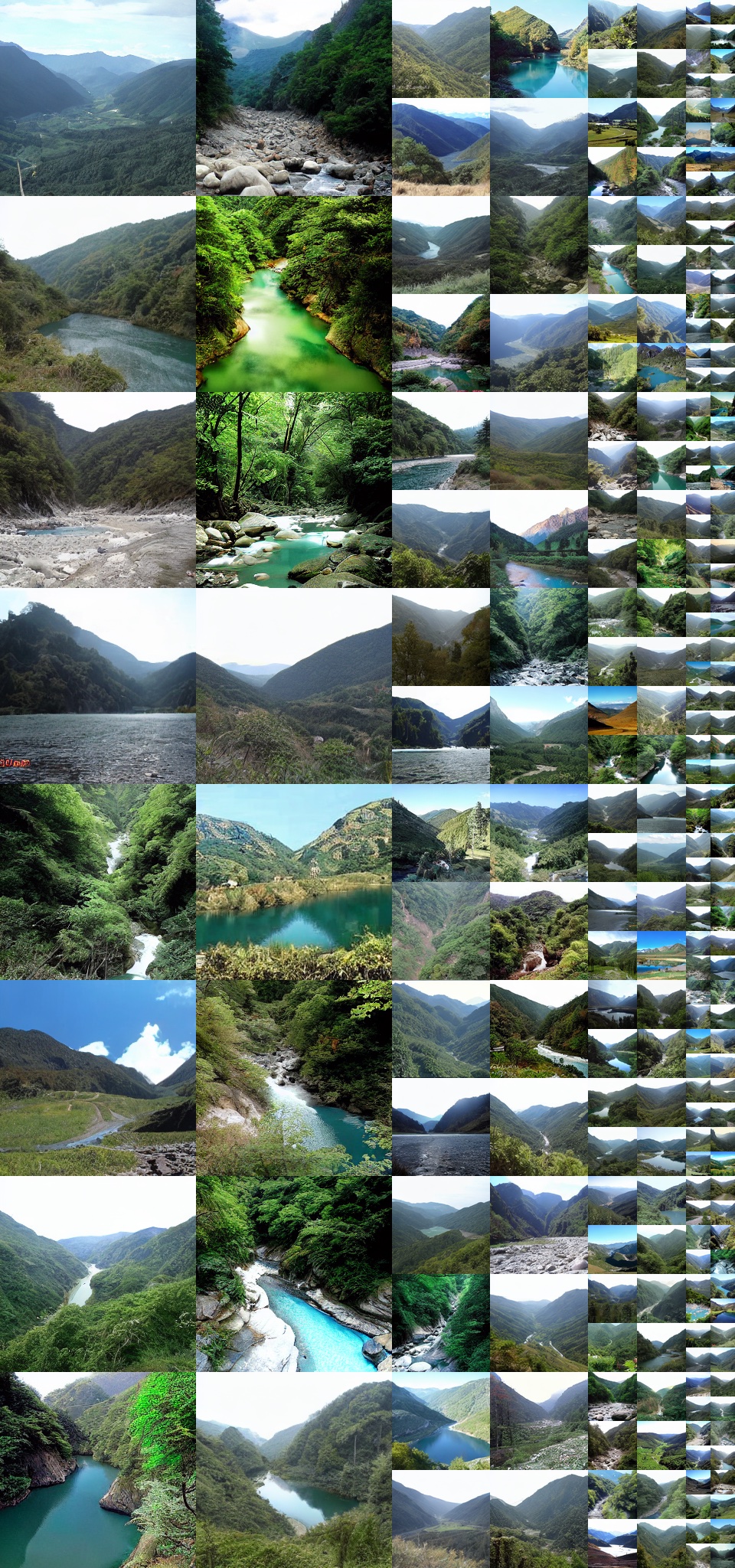}
      \caption[]{\textbf{Uncurated $256\times256$ SiT-XL samples}.\\
           Classifier-free guidance scale = 4.0 \\
           Class label = "valley"(979)
        }
        \label{fig:superimage-979}
  \end{minipage}
\end{figure}

\newpage 
\end{document}